\documentclass[journal,twoside]{IEEEtran}
\usepackage{amsmath, amssymb, amsfonts}
\usepackage{url}
\usepackage{marvosym}
\usepackage{balance}
\usepackage[table,xcdraw]{xcolor}
\usepackage{cite}
\usepackage[hidelinks,hypertexnames=false,colorlinks=true,linkcolor=blue!80!black,citecolor=red!70!black,urlcolor=black]{hyperref}
\usepackage{graphicx}
\usepackage{booktabs}
\usepackage{multirow}
\usepackage{threeparttable}
\usepackage{makecell}
\usepackage{mathrsfs}
\usepackage{enumitem}
\usepackage{ulem}
\usepackage{float}
\usepackage{orcidlink}
\usepackage{amsmath}
\usepackage{array}
\usepackage{amsthm}
\usepackage{mathtools}
\usepackage{xurl}
\usepackage{dsfont}
\usepackage{cuted}
\usepackage{capt-of}

\usepackage[switch]{lineno}
\usepackage[table,xcdraw]{xcolor}
\usepackage{scalerel}
\usepackage{tikz}
\usetikzlibrary{svg.path}

\definecolor{orcidlogocol}{HTML}{A6CE39}
\tikzset{
  orcidlogo/.pic={
    \fill[orcidlogocol] svg{M256,128c0,70.7-57.3,128-128,128C57.3,256,0,198.7,0,128C0,57.3,57.3,0,128,0C198.7,0,256,57.3,256,128z};
    \fill[white] svg{M86.3,186.2H70.9V79.1h15.4v48.4V186.2z}
                 svg{M108.9,79.1h41.6c39.6,0,57,28.3,57,53.6c0,27.5-21.5,53.6-56.8,53.6h-41.8V79.1z M124.3,172.4h24.5c34.9,0,42.9-26.5,42.9-39.7c0-21.5-13.7-39.7-43.7-39.7h-23.7V172.4z}
                 svg{M88.7,56.8c0,5.5-4.5,10.1-10.1,10.1c-5.6,0-10.1-4.6-10.1-10.1c0-5.6,4.5-10.1,10.1-10.1C84.2,46.7,88.7,51.3,88.7,56.8z};
  }
}
\newcommand\orcidicon[1]{\href{https://orcid.org/#1}{\mbox{\scalerel*{
\begin{tikzpicture}[yscale=-1,transform shape]
\pic{orcidlogo};
\end{tikzpicture}
}{|}}}}
\newcolumntype{L}[1]{>{\raggedright\arraybackslash}p{#1}}
\newcolumntype{C}[1]{>{\centering\arraybackslash}p{#1}}
\newcolumntype{R}[1]{>{\raggedleft\arraybackslash}p{#1}}

\definecolor{TYS}{rgb}{0.6, 0.8, 0.2}
\definecolor{DOcolor}{rgb}{1,0.45,0.0}
\definecolor{NAVYcolor}{rgb}{0.05,0,0.5}
\definecolor{rowlight}{RGB}{240,240,240}
\definecolor{postgreen}{RGB}{214,236,214}
\definecolor{noliyellow}{RGB}{255,246,200}
\definecolor{headblue}{RGB}{213,232,252}

\newcommand{\settablefont}{\fontsize{7.5}{10}\selectfont}
\newcommand{\settablecitationfont}{\fontsize{6}{8}\selectfont}
\newcommand{\egi}{\textit{e.g.}}

\newcommand{\eg}{\egi}

\newcommand\clgray[1]{{\color{gray}{#1}}}

\usepackage{titlecaps}
\usepackage{arydshln}

\Addlcwords{a an and as at by for from in of on or the to with}

\begin{document}

\title{\huge
Learning Optical Flow with Task-Oriented Feature Adaptation
}

\author{
    Mengtan~Zhang\,\orcidlink{0009-0003-3468-7680},
    Dazhong~Wang\,\orcidlink{0009-0001-8941-6234},
    Haoran~Zhang\,\orcidlink{0009-0009-7578-5897},
    Hongbo~Zhao\,\orcidlink{0009-0008-2198-2484},
    Yuhang~Ming\,\orcidlink{0000-0002-4548-6388},
    Weihao~Yuan\,\orcidlink{0000-0002-1362-3747},
    Wei~Ye\,\orcidlink{0000-0002-3784-7788},~\IEEEmembership{Member,~IEEE},
    Hanli~Wang\,\orcidlink{0000-0002-9999-4871},~\IEEEmembership{Senior Member,~IEEE},
    and Rui~Fan\,\orcidlink{0000-0003-2593-6596},~\IEEEmembership{Senior Member,~IEEE}
    \thanks{
    \textit{(Mengtan Zhang and Dazhong Wang contributed equally to this work.)}
    \textit{(Corresponding author: Rui Fan.)}
    }
    \thanks{Mengtan Zhang, Dazhong Wang and Hongbo Zhao are with the Shanghai Research Institute for Intelligent Autonomous Systems, Tongji University, Shanghai 201210, P. R. China (e-mail: \{zmt200110, 2353106, hongbozhao\}@tongji.edu.cn).}
    \thanks{Haoran Zhang and Wei Ye are with the College of {Electronic} \& Information Engineering, Tongji University, Shanghai {201804}, P. R. China (e-mail: \{2251736, yew\}@tongji.edu.cn).}
    \thanks{Yuhang Ming is with the School of Computer Science, Hangzhou Dianzi University, Hangzhou 310018, China (e-mail: yuhang.ming@hdu.edu.cn).}
    \thanks{Weihao Yuan is with the School of Intelligence Science and Technology, Nanjing University, Suzhou 215163, China (e-mail: dr.weihao.yuan@gmail.com).}
    \thanks{Hanli Wang is with the College of {Electronic} and Information Engineering, the School of Computer Science and Technology, and the Key Laboratory of Embedded System and Service Computing (Ministry of Education), Tongji University, Shanghai {201804}, P. R. China (e-mail: hanliwang@tongji.edu.cn).}
    \thanks{
    {Rui Fan is with the College of Electronic and Information Engineering, Shanghai Institute of Intelligent Science and Technology, Shanghai Research Institute for Intelligent Autonomous Systems, Shanghai Key Laboratory of Intelligent Autonomous Systems, Urban  Mobility  Institute, State Key Laboratory of Autonomous Intelligent Unmanned Systems, and Frontiers Science Center for Intelligent Autonomous Systems (Ministry of Education), The Key Laboratory of Road and Traffic Engineering (Ministry of Education), Tongji University, Shanghai 201804, China, as well as with the National Key Laboratory of Human-Machine Hybrid Augmented Intelligence, Xi'an Jiaotong University, Xi'An, Shaanxi 710049, China (e-mail: rui.fan@ieee.org).}
    }
}

\maketitle

\begin{abstract}
Learning-based optical flow estimation approaches generally rely on both inter-frame correspondence evidence and intra-frame contextual guidance, yet features from general-purpose image encoders may not fully support these roles. Strongly activated but weakly discriminative channels can outweigh more informative ones when computing inter-frame feature similarity, while context feature affinities can be low within motion-coherent regions yet high across motion boundaries. To address these mismatches, this study proposes TO-Flow, a task-oriented feature adaptation framework that aligns matching and context features with their respective requirements through dedicated channel modulation and spatial reorganization approaches. An affinity-based soft partition distinguishes motion-region interiors from boundary-adjacent regions to guide both processes. Channel modulation uses simulated flow refinement to assess whether each channel assigns higher similarity to estimated correspondences than to nearby competing candidates. The resulting similarity differences are aggregated within each region type and averaged across iterations to derive channel weights that emphasize informative channels and suppress ambiguous ones. Spatial reorganization combines affinity-guided feature attraction and repulsion from coarse to fine to simultaneously improve consistency within motion-coherent regions and distinctiveness across motion boundaries, with feature resampling mitigating boundary blurring during cross-scale fusion.
Extensive experiments on the MPI Sintel, KITTI 2015, Spring, and LayeredFlow datasets demonstrate the effectiveness and general applicability of TO-Flow across diverse iterative refinement architectures as well as supervised and unsupervised training paradigms.
The source code will be publicly available at \url{https://mias.group/TO-Flow} upon publication.

\end{abstract}


\begin{figure*}[t]
	\centering
	\includegraphics[width=0.99\textwidth]{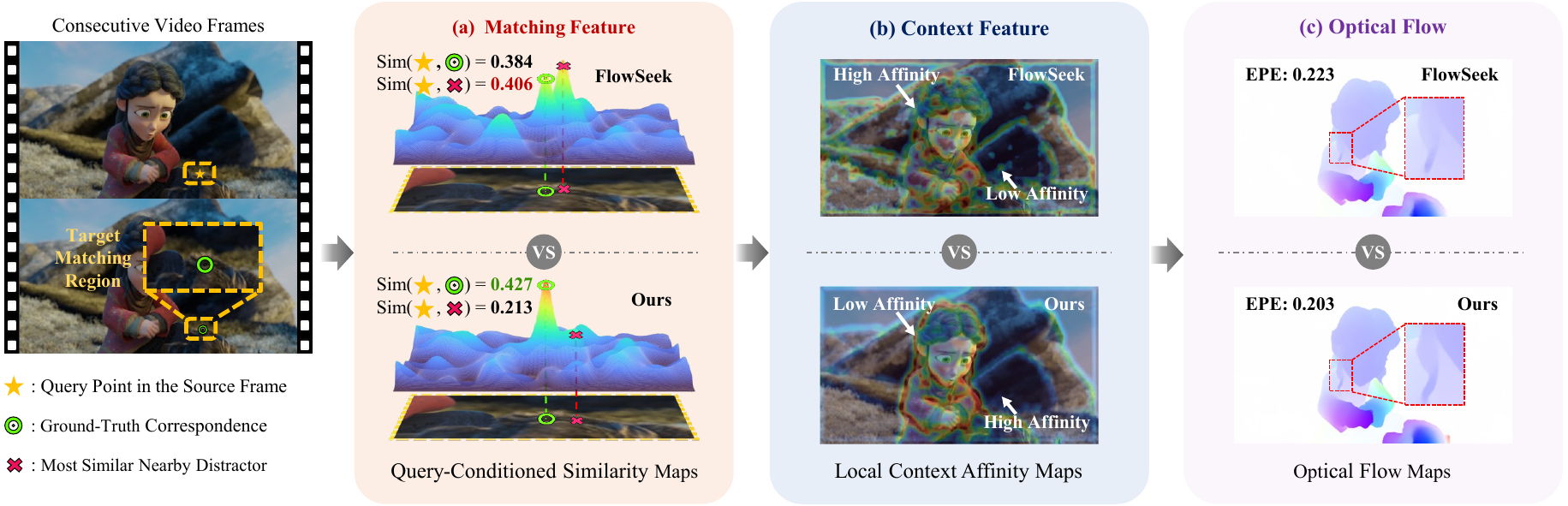}
	\caption{
		Compared with the recurrent refinement baseline FlowSeek-T, TO-Flow provides (a) more discriminative matching features, with higher query similarity to the ground-truth correspondence than to nearby distractors, and (b) more consistent context features, with higher local affinity within motion-coherent regions and lower affinity near motion boundaries. These improvements support (c) more accurate optical flow estimation.
	}
	\label{fig:motivation}
\end{figure*}

\section{Introduction}\label{sec:introduction}

\IEEEPARstart{O}{ptical} flow estimation is a fundamental low-level vision task that aims to recover dense pixel-wise motion between consecutive images, providing motion cues for a broad range of downstream vision applications including geometric perception \cite{bian2021unsupervised, luo2020consistent, chi2022subspace, zhang2024dcpi, shao2025iebins, zhang2026discriminately}, dynamic scene reconstruction \cite{sun2025splatflow,zheng2025flow,feng2026panobench}, video understanding and generation \cite{jin2025flovd, simonyan2014two}, and embodied world modeling \cite{bi2026motus,chen2026flowwam, zhang2026skytopia}.

Modern learning-based approaches \cite{sun2018pwc,teed2020raft,jiang2021learning,huang2022flowformer,xu2022gmflow,luo2024flowdiffuser,luo2026optical} typically infer optical flow by jointly exploiting inter-frame correspondence cues and intra-frame contextual information.
In particular, the success of RAFT \cite{teed2020raft} established recurrent refinement as a dominant paradigm for optical flow estimation \cite{wang2024sea,poggi2025flowseek,wang2026waft}. Under this paradigm, an update operator is employed to iteratively reduce residual flow errors by leveraging both inter-frame correspondence evidence and intra-frame contextual guidance. Existing methods mainly differ in how the correspondence evidence is obtained, with one type explicitly querying feature correlation volumes \cite{teed2020raft,wang2024sea,poggi2025flowseek} and the other progressively aligning cross-frame features through flow-guided warping \cite{wang2026waft}. Meanwhile, the contextual guidance is typically obtained by directly using features extracted from a standard convolutional encoder \cite{he2016deep}.
Regardless of the refinement process, the reliability of recurrent updates ultimately depends on the quality of both correspondence and context features. Specifically, matching features require sufficient discriminability for accurate inter-frame matching, while context features should provide informative cues for propagating reliable motion information and promoting spatially consistent flow refinement. However, less attention has been paid to whether features extracted by general-purpose image encoders are well suited for direct use in optical flow estimation.

As demonstrated in Fig.~\ref{fig:motivation}(a), this study first investigates matching features by visualizing a feature similarity map over the target frame for a query point in the source frame. It can be observed that even matching features produced by a highly competitive optical flow model \cite{poggi2025flowseek} can exhibit incorrect responses, with neighboring distractors showing higher similarity than the ground-truth correspondence. This observation reflects that existing matching features may contain redundant or task-irrelevant channels that are sensitive to inter-frame appearance changes, illumination variations, viewpoint shifts, and occlusions, preventing the resulting representations from encoding sufficiently stable and discriminative correspondence information. Such ambiguous matching features can therefore introduce unreliable matching cues into the iterative update process and ultimately degrade optical flow estimation.

In parallel with the matching features, this study further visualizes the local affinity map derived from the context features in Fig.~\ref{fig:motivation}(b). Specifically, for each pixel, its affinity score is computed as the average cosine similarity between its feature and those of the surrounding pixels. Ideally, pixels within a motion-coherent region should exhibit high affinity, reflecting consistent contextual representations, whereas pixels near motion boundaries should show lower affinity to distinguish neighboring regions with different motions. However, existing context features do not consistently follow this expected pattern, with high affinity appearing across motion boundaries and low affinity occurring within otherwise motion-coherent regions. Such misleading contextual cues can in turn affect iterative flow updates, leading to erroneous propagation across motion boundaries.

To address the aforementioned limitations, this study proposes \textbf{TO-Flow}, a \textbf{T}ask-\textbf{O}riented feature adaptation framework for optical \textbf{Flow} estimation that adapts front-end features to their distinct roles in providing correspondence evidence and contextual guidance, rather than directly reusing them as generic visual representations.

Specifically, TO-Flow first derives a feature adaptation partition mask that coarsely separates motion-coherent regions from boundary-adjacent regions, as these two types of locations require different adaptation behaviors. Within coherent regions, matching features should emphasize channels that better distinguish the current match from local alternatives, while context features should encourage consistent motion propagation. Near motion boundaries, matching features should suppress channels that become unreliable under local variations, while context features should preserve sharp discontinuities to avoid cross-boundary propagation. Building on this region-aware partition, TO-Flow adapts the matching and context features along the channel and spatial dimensions, respectively.

Matching features are processed by channel modulation from simulated matching feedback, which assesses and adjusts individual feature channels according to their matching behavior during flow refinement. A short stop-gradient refinement simulation that follows the original iterative refinement process is first performed. At each simulated iteration, the current flow estimate defines a correspondence hypothesis for each source-frame pixel, and the channel-wise response at this hypothesis is compared with those of nearby competing candidates. This comparison provides task-specific feedback on whether each channel increases or decreases the response margin between the current match and its local alternatives. This feedback is aggregated separately within the aforementioned partitioned regions and averaged across simulated iterations to derive channel reweighting vectors. The resulting vectors modulate the corresponding channels within each region type, enhancing channels that consistently provide distinctive matching evidence while suppressing those that respond ambiguously to competing candidates. In this way, progressively refined flow estimates provide increasingly informative task feedback for identifying feature channels that support reliable correspondence estimation, while the adapted features, in turn, facilitate more accurate subsequent flow refinement.

Context features are refined through spatial reorganization using an attraction-and-repulsion mechanism to simultaneously improve feature consistency within the interiors of motion-coherent regions and increase feature distinctiveness across motion boundaries. Two affinity-guided adaptive convolution kernels are constructed and applied to boundary-adjacent and motion-coherent regions, respectively.
Near motion boundaries, the first kernel assigns larger weights to less similar neighboring locations based on local feature affinity and is used to aggregate features from the opposite motion region. The aggregated feature is then compared with the original feature, and their difference is used to further enlarge the representation discrepancy between the two sides of the boundary.
Meanwhile, within motion-coherent regions, the second kernel assigns larger weights to more similar neighboring locations and is used to aggregate neighboring features within the same motion region, promoting consistent feature propagation throughout the region.
These two operations are applied alternately from coarse to fine, progressively enabling the context features to represent each motion-coherent region as a whole, thereby providing more reliable contextual guidance for subsequent iterative flow refinement.

TO-Flow can be combined seamlessly with diverse optical flow architectures and applied under both supervised and unsupervised training paradigms.
Extensive experiments on the MPI Sintel, KITTI 2015, Spring, and LayeredFlow datasets demonstrate that incorporating TO-Flow improves representative supervised optical flow models with different iterative refinement mechanisms, including FlowSeek~\cite{poggi2025flowseek} and WAFT~\cite{wang2026waft}, which refine optical flow using a cost volume and feature warping, respectively. Improvements are also observed when TO-Flow is integrated into the unsupervised U$^{2}$Flow framework~\cite{sun2026u}.
These results demonstrate the effectiveness of task-oriented feature adaptation and its general applicability for improving optical flow estimation.
In summary, the main contributions of this work are as follows:

\begin{itemize}
	\item A phenomenological analysis of the mismatch between front-end feature representations and the distinct requirements of correspondence matching and contextual guidance, based on which task-oriented feature adaptation is proposed as a new perspective for addressing this mismatch.

	\item Channel modulation from simulated matching feedback, which assesses and adjusts individual feature channels according to their matching behavior during flow refinement, enhancing channels that provide distinctive matching evidence while suppressing those that respond ambiguously to competing candidates.

	\item Spatial reorganization using an attraction-and-repulsion mechanism, which improves feature consistency within motion-coherent regions and increases feature distinctiveness across motion boundaries through coarse-to-fine processing, providing more reliable contextual guidance for subsequent iterative flow refinement.

	\item Extensive experiments on multiple standard optical flow evaluation benchmarks validate TO-Flow's effectiveness and general applicability across different iterative refinement backends and both supervised and unsupervised training paradigms.

\end{itemize}

The remainder of this article is organized as follows: Sect.~\ref{sec:related_work} reviews the development of optical flow estimation, as well as existing channel modulation and spatial reorganization methods. Sect.~\ref{sec:method} presents the proposed TO-Flow framework. Sect.~\ref{sec:experiments} presents the experimental results across several public datasets. Finally, Sect.~\ref{sec:conclusion} concludes this article and discusses the research directions for future work.

\section{Related Work}
\label{sec:related_work}
\subsection{Optical Flow Estimation}

With advances in deep learning, optical flow estimation has gradually shifted from traditional methods \cite{horn1981determining,black1993framework,zach2007duality,bruhn2005lucas,papenberg2006highly} that formulate it as an optimization problem to deep learning approaches \cite{sun2018pwc,hui2018liteflownet,teed2020raft,li2024gyroflow+}. Early learning-based methods \cite{sun2018pwc,hui2018liteflownet} typically extract feature pyramids from each input frame and compute local feature correlations at each scale to estimate optical flow through coarse-to-fine refinement. However, this strategy can miss small, fast-moving objects and struggle to correct errors introduced at coarse resolutions \cite{teed2020raft}. In contrast, RAFT \cite{teed2020raft} iteratively refines a single high-resolution optical flow map using multi-scale 4D correlation volumes constructed from all pairs of matching features, with context features guiding the updates.

Building upon RAFT's iterative refinement framework, numerous subsequent methods \cite{sui2022craft,jahedi2024ms,wang2024splatflow} have been proposed. CRAFT \cite{sui2022craft} introduces feature transformation and cross-frame attention to compute more accurate correlations. MS-RAFT+ \cite{jahedi2024ms} combines recurrent refinement with coarse-to-fine estimation to resolve matching ambiguities and recover fine spatial details. More recently, optical flow estimation has also been explored from a generative perspective \cite{saxena2023surprising,luo2024flowdiffuser}. FlowDiffuser \cite{luo2024flowdiffuser} progressively denoises a randomly initialized flow map using matching and context cues.

Despite differences in their estimation strategies, matching and context features play a fundamental role in these frameworks, motivating this work to refine both feature representations.

\subsection{Channel Modulation}

Early methods~\cite{hu2018squeeze,woo2018cbam} use global pooling to construct channel descriptors, which are then used to predict weights for channel modulation.
However, channels with different spatial response distributions may produce similar descriptors after such pooling. Other methods therefore incorporate richer information when deriving channel weights. GSoP-Net \cite{gao2019global} further computes covariance between pairs of channels across spatial locations, while MCA \cite{jiang2024mca} computes the mean and variance of spatial responses within each channel and fuses them through convolution.

Nevertheless, strong responses within a frame do not necessarily reflect a channel's ability to distinguish corresponding locations from competing matches across frames.
TO-Flow instead uses simulated flow refinement to assess how well individual channels distinguish estimated correspondences from nearby competing candidates and modulates their contributions accordingly.

\subsection{Spatial Reorganization}

Typical methods~\cite{hu2019local,li2021involution,su2019pixel} refine spatial representations through adaptive aggregation of neighboring features. However, such local aggregation may be insufficient to distinguish regions with similar feature representations. Other methods are therefore introduced to incorporate information from broader spatial regions. LSKNet \cite{li2025lsknet} uses convolutional kernels of different sizes to aggregate features beyond local neighborhoods, while LiSA \cite{kwon2025lightweight} incorporates relative positional information into attention to guide feature aggregation across the whole image.

Nevertheless, applying a uniform feature reorganization strategy across the entire feature map may mix features from different motion regions, introducing misleading cues for optical flow estimation. In contrast, TO-Flow selectively applies feature attraction within motion-coherent regions and feature repulsion near motion boundaries to simultaneously promote intra-region consistency and strengthen distinctions between adjacent regions.

\begin{figure*}[!t]
	\centering
	\includegraphics[width=0.99\textwidth]{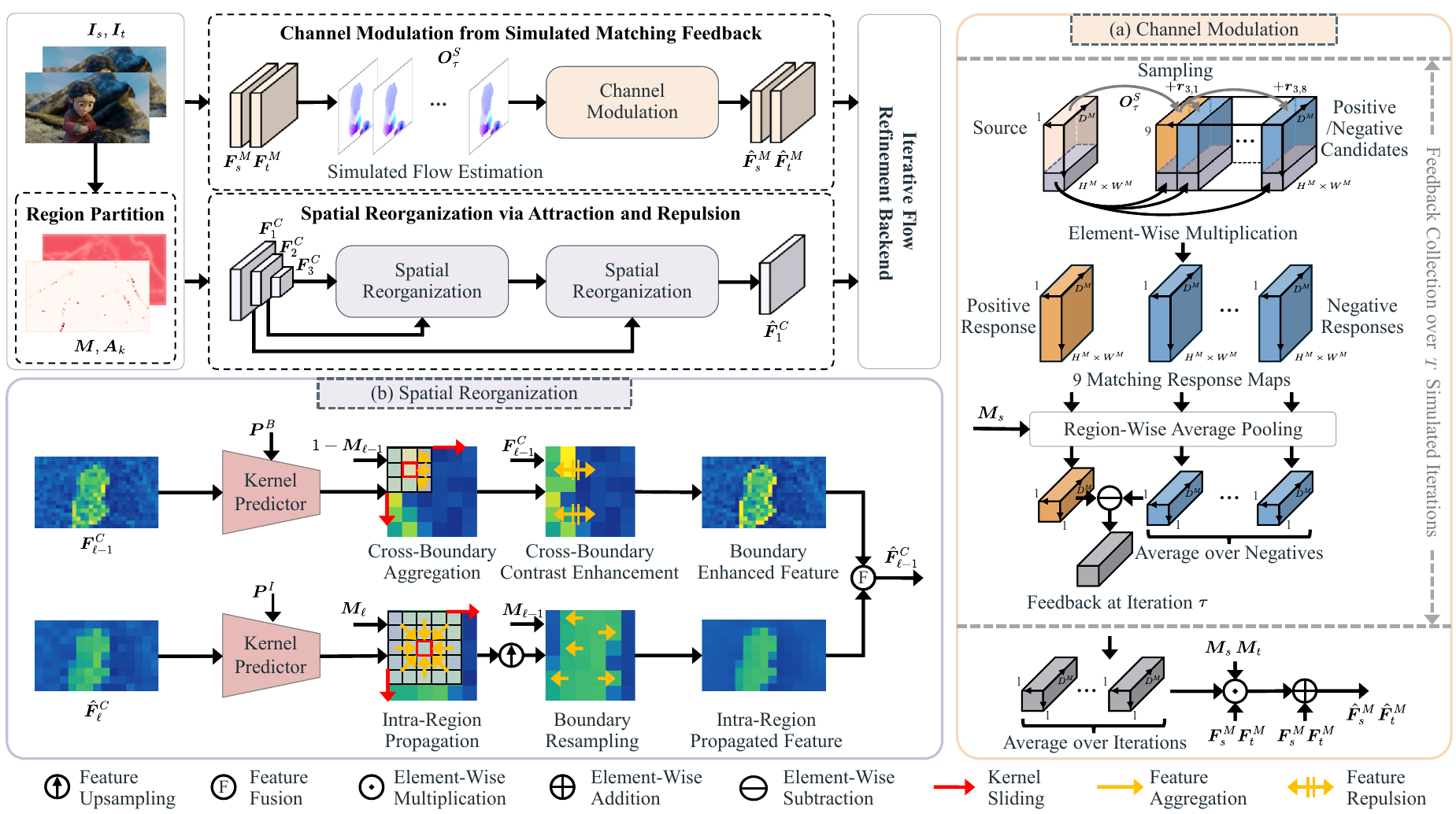}
	\caption{
		Overview of the TO-Flow framework. An affinity-based soft region partition guides two processes before iterative flow refinement: (a) channel modulation of matching features using simulated matching feedback and (b) spatial reorganization of context features through cross-boundary contrast enhancement and intra-region propagation across resolutions.
	}
	\label{framework}
\end{figure*}

\section{Method}
\label{sec:method}

\subsection{Framework Overview}
\label{3.1}
The overall framework of TO-Flow is illustrated in Fig.~\ref{framework}.
Given consecutive source and target frames $\boldsymbol{I}_{s},\boldsymbol{I}_{t}\in\mathbb{R}^{H\times W\times3}$ as inputs, where $H$ and $W$ denote the image height and width, respectively, two matching feature maps $\boldsymbol{F}_{s}^{M},\boldsymbol{F}_{t}^{M}\in\mathbb{R}^{H^{M}\times W^{M}\times D^{M}}$ and a context feature map $\boldsymbol{F}^{C}\in\mathbb{R}^{H^{C}\times W^{C}\times D^{C}}$ are extracted by two separate encoders, respectively, with their channel dimensions denoted by $D^{M}$ and $D^{C}$, their heights by $H^{M}$ and $H^{C}$, and their widths by $W^{M}$ and $W^{C}$\footnote{As a general plug-and-play feature adaptation approach compatible with different optical flow networks, TO-Flow preserves each network's feature map resolutions and channel dimensions.}.
Each extracted feature map is then processed by the affinity-based soft region partitioning approach described in Sect.~\ref{3.2}, yielding a soft partition that distinguishes boundary-adjacent regions from the interiors of motion-coherent regions. The resulting partitions guide the two subsequent feature adaptation processes.
For matching features, channel modulation based on simulated matching feedback derives two region-specific channel reweighting vectors from matching responses collected over simulated flow refinement iterations performed without gradient backpropagation and uses these vectors to modulate the matching features in the corresponding regions, as detailed in Sect.~\ref{3.3}.
For context features, spatial reorganization via attraction and repulsion proceeds from coarse to fine, using region-specific aggregation kernels generated from context features and local affinities to enhance feature differences across motion boundaries and improve feature consistency within motion-coherent regions, as detailed in Sect.~\ref{3.4}.
Finally, the adapted matching and context features are fed into the recurrent refinement backend to iteratively estimate the optical flow map $\boldsymbol{O}\in\mathbb{R}^{H\times W\times2}$ from $\boldsymbol{I}_{s}$ to $\boldsymbol{I}_{t}$.

\subsection{Affinity-Based Soft Region Partition}
\label{3.2}

As discussed in Sect.~\ref{sec:introduction}, motion-coherent regions and boundary-adjacent regions impose different requirements on feature adaptation, motivating a region-dependent feature adaptation strategy. A key challenge is therefore to identify these two types of regions directly from the input features, without requiring explicit region type annotations. A natural approach is to partition regions based on local feature affinity, treating low affinity as an indicator of motion boundaries and high affinity as a cue for motion-coherent regions. However, as Fig.~\ref{fig:motivation} illustrates, context features extracted by a generic encoder do not always provide reliable local affinity cues. In particular, feature noise can produce low affinity within motion-coherent regions, causing these locations to be mistakenly classified as boundary-adjacent regions.
This introduces a circular dependency, as region-dependent feature adaptation requires a reliable partition, while obtaining such a partition requires feature adaptation to suppress noise in the local affinities.
To address this issue, this study proposes a soft region partitioning strategy that first suppresses isolated feature noise based on the maximum similarity within each local neighborhood and then derives soft region assignments from the refined local affinities.

For each matching or context feature map $\boldsymbol{F}\in\{\boldsymbol{F}_{s}^{M}, \boldsymbol{F}_{t}^{M},\boldsymbol{F}^{C}\}$, consider a $k\times k$ neighborhood centered at pixel $\boldsymbol{p}=(u,v)^{\top}$. Let $\mathcal{N}_{k}=\{\boldsymbol{r}_{k,j}\}_{j=1}^{k^{2}-1}\subset\mathbb{Z}^{2}$ denote the set of offsets from $\boldsymbol{p}$ to the other $(k^{2}-1)$ pixels in this neighborhood, with $\boldsymbol{r}_{k,0}=(0,0)^{\top}$ denoting the center offset. For $\boldsymbol{r}\in\mathcal{N}_{k}$, the cosine similarity between the feature vectors $\boldsymbol{F}(\boldsymbol{p})$ and $\boldsymbol{F}(\boldsymbol{p}+\boldsymbol{r})$ is computed as follows:
\begin{equation}
	s(\boldsymbol{p},\boldsymbol{r})
	=\frac{\boldsymbol{F}(\boldsymbol{p})^{\top}
		\boldsymbol{F}(\boldsymbol{p}+\boldsymbol{r})}
	{\|\boldsymbol{F}(\boldsymbol{p})\|_{2}\,
		\|\boldsymbol{F}(\boldsymbol{p}+\boldsymbol{r})\|_{2}},
	\label{eq:local_feature_similarity}
\end{equation}
where $\|\cdot\|_{2}$ denotes the Euclidean norm.
A key distinction between boundary-adjacent features and isolated noisy features is that the former can receive local support from similar neighboring features on the same side of the boundary, whereas the latter may differ from all their neighbors and thus lack such support.
The maximum similarity within a $3\times3$ neighborhood is therefore used to measure such local feature support as follows:
\begin{equation}
	h(\boldsymbol{p})
	=\max_{\boldsymbol{r}\in\mathcal{N}_{3}}
	s(\boldsymbol{p},\boldsymbol{r}).
	\label{eq:local_feature_support}
\end{equation}
A lower support score $h(\boldsymbol{p})$ indicates weaker support from neighboring features, suggesting a greater need for smoothing.
Accordingly, greater weight is assigned to the local neighborhood average when combining it with the original feature to yield the smoothed feature $\widetilde{\boldsymbol{F}}(\boldsymbol{p})$ as follows:
\begin{equation}
	\widetilde{\boldsymbol{F}}(\boldsymbol{p})
	=\left(1-\eta(\boldsymbol{p})\right)\boldsymbol{F}(\boldsymbol{p})+\eta(\boldsymbol{p})\bar{\boldsymbol{F}}(\boldsymbol{p}),
	\label{eq:noise_aware_smoothing}
\end{equation}
where $\eta(\boldsymbol{p})=\sigma(\theta-h(\boldsymbol{p}))$ denotes the adaptive smoothing weight, $\theta$ is a threshold such that the neighborhood average receives greater weight than the original feature when $h(\boldsymbol{p})<\theta$, $\bar{\boldsymbol{F}}(\boldsymbol{p})$ denotes the average of $\boldsymbol{F}(\boldsymbol{p}+\boldsymbol{r})$ over $\boldsymbol{r}\in\mathcal{N}_{3}\cup\{\boldsymbol{r}_{3,0}\}$, and $\sigma(\cdot)$ denotes the sigmoid function.
Replacing $\boldsymbol{F}$ with $\widetilde{\boldsymbol{F}}$ in \eqref{eq:local_feature_similarity} and computing the similarities over all pixels and neighborhood offsets yields a refined similarity map $\widetilde{\boldsymbol{S}}_{k}$ of size $H^{M}\times W^{M}\times(k^{2}-1)$ for matching features or $H^{C}\times W^{C}\times(k^{2}-1)$ for context features. Applying the sigmoid function to $\widetilde{\boldsymbol{S}}_{k}$ then yields an affinity map $\boldsymbol{A}_{k}=\sigma(\widetilde{\boldsymbol{S}}_{k})$.

After noise suppression, low affinity to even one neighboring pixel may indicate that the local neighborhood crosses a motion boundary.
A soft mask $\boldsymbol{M}$ is therefore computed from the minimum of the eight affinity scores at each pixel in $\boldsymbol{A}_{3}$ as follows:
\begin{equation}
	\boldsymbol{M}(\boldsymbol{p})
	=\sigma\!\left(
	\min_{j=1,\ldots,8}\boldsymbol{A}_{3}(\boldsymbol{p},j)
	+\theta_{\mathrm{b}}-1
	\right),
	\label{eq:motion_region_partition}
\end{equation}
where $\boldsymbol{A}_{3}(\boldsymbol{p},j)$ denotes the affinity between pixel $\boldsymbol{p}$ and its $j$-th neighboring pixel, and $\theta_{\mathrm{b}}$ denotes the boundary threshold.
The resulting mask, of size $H^{M}\times W^{M}$ for matching features or $H^{C}\times W^{C}$ for context features, assigns an interior weight $\boldsymbol{M}(\boldsymbol{p})\in(0,1)$ and a boundary-adjacent weight $1-\boldsymbol{M}(\boldsymbol{p})$ to each pixel.

The partition is computed separately for each feature map at each frame and scale.
The masks guide the modulation of the original matching features in Sect.~\ref{3.3} and, together with the local affinities, the spatial reorganization of the original context features in Sect.~\ref{3.4}.

\subsection{Channel Modulation from Simulated Matching Feedback}
\label{3.3}

The similarity score for correspondence matching between two pixels is computed as the inner product of their feature vectors, aggregating contributions across channels. The matching ambiguity in Fig.~\ref{fig:motivation}(a) may therefore arise when channels with strong activations but limited ability to distinguish true correspondences from distractors dominate the similarity score, while more discriminative channels contribute less because of their weaker activations. Moreover, although feature channels are shared across the image, the cues important for correspondence matching may differ between boundary-adjacent regions and the interiors of motion-coherent regions. This mismatch between the activation strength and matching discriminability of channels may therefore vary across the two region types. To address this issue, TO-Flow adapts the matching features using channel modulation from simulated matching feedback. This approach uses a preliminary flow refinement simulation to generate correspondence hypotheses and derive channel reweighting vectors as feedback from local matching competition, without requiring ground-truth correspondences. To obtain these vectors, differences in channel responses between the hypotheses and neighboring alternatives are aggregated separately within the two region types and averaged across iterations, reducing dependence on any single, potentially inaccurate hypothesis. The two vectors then adaptively adjust channel contributions during the forward pass to improve the discriminability of matching features.

Specifically, TO-Flow first uses the recurrent backend to perform $T$ simulated flow refinement iterations with the original matching and context features, producing flow maps $\boldsymbol{O}_{\tau}^{S}\in\mathbb{R}^{H^{M}\times W^{M}\times2}$ for $\tau=1,\ldots,T$. The flow displacements are measured in pixels of the matching feature maps.
At each simulated iteration, $\boldsymbol{O}_{\tau}^{S}$ maps each pixel in the source matching feature map to a provisional correspondence in the target matching feature map.
Sampling the target matching feature map at these correspondences and their eight neighboring pixels yields nine warped feature maps $\boldsymbol{F}_{\tau,j}^{W}\in\mathbb{R}^{H^{M}\times W^{M}\times D^{M}}$ for $j=0,\ldots,8$, as follows:
\begin{equation}
	\boldsymbol{F}_{\tau,j}^{W}(\boldsymbol{p})
	=\mathcal{W}\!\left(
	\boldsymbol{F}_{t}^{M},
	\boldsymbol{p}+\boldsymbol{O}_{\tau}^{S}(\boldsymbol{p})+\boldsymbol{r}_{3,j}
	\right),
	\label{eq:feature_warp}
\end{equation}
where $\mathcal{W}(\cdot,\cdot)$ denotes the feature warping function based on bilinear interpolation.
The center sample ($j=0$) represents a hypothesized positive, and the eight neighboring samples represent negative alternatives.
Subsequently, element-wise multiplication of the source matching feature map and each warped feature map yields the matching response map $\boldsymbol{R}_{\tau,j}=\boldsymbol{F}_{s}^{M}\odot\boldsymbol{F}_{\tau,j}^{W}$, retaining the matching response of each channel, where $\odot$ denotes element-wise multiplication. Using the mask $\boldsymbol{M}_{s}$ obtained from $\boldsymbol{F}_{s}^{M}$ through the partitioning process in Sect.~\ref{3.2}, these responses are averaged separately over the interiors of motion-coherent regions and boundary-adjacent regions as follows:
\begin{equation}
	\boldsymbol{s}_{\tau,j}^{I}
	=\frac{\sum_{\boldsymbol{p}}\boldsymbol{M}_{s}(\boldsymbol{p})
		\boldsymbol{R}_{\tau,j}(\boldsymbol{p})}
	{\sum_{\boldsymbol{p}}\boldsymbol{M}_{s}(\boldsymbol{p})},
	\label{eq:region_matching_response}
\end{equation}
and
\begin{equation}
	\boldsymbol{s}_{\tau,j}^{B}
	=\frac{\sum_{\boldsymbol{p}}\left(1-\boldsymbol{M}_{s}(\boldsymbol{p})\right)
		\boldsymbol{R}_{\tau,j}(\boldsymbol{p})}
	{\sum_{\boldsymbol{p}}\left(1-\boldsymbol{M}_{s}(\boldsymbol{p})\right)},
	\label{eq:boundary_matching_response}
\end{equation}
where $\boldsymbol{s}_{\tau,j}^{I},\boldsymbol{s}_{\tau,j}^{B} \in\mathbb{R}^{D^{M}}$ denote the weighted average matching responses of individual channels within the interiors of motion-coherent regions and boundary-adjacent regions, respectively.
The difference between the positive response and the mean negative response is averaged over the simulated iterations and then fed into two independent multilayer perceptrons (MLPs) to produce two channel reweighting vectors $\boldsymbol{w}^{G}\in\mathbb{R}^{D^{M}}$ as follows:
\begin{equation}
	\boldsymbol{w}^{G}
	=\operatorname{MLP}^{G}\left(\frac{1}{T}\sum_{\tau=1}^{T}
	\left(\boldsymbol{s}_{\tau,0}^{G}
	-\frac{1}{8}\sum_{j=1}^{8}\boldsymbol{s}_{\tau,j}^{G}\right)\right),
	\label{eq:channel_feedback}
\end{equation}
where $G\in\{I,B\}$ denotes the region type.
These two vectors are combined using the mask $\boldsymbol{M}_{i}$ of $\boldsymbol{F}_{i}^{M}$ to obtain a channel-wise modulation map $\boldsymbol{\Gamma}_{i}$ for each frame $i\in\{s,t\}$ as follows:
\begin{equation}
	\boldsymbol{\Gamma}_{i}(\boldsymbol{p})
	=\boldsymbol{M}_{i}(\boldsymbol{p})\boldsymbol{w}^{I}+\left(1-\boldsymbol{M}_{i}(\boldsymbol{p})\right)\boldsymbol{w}^{B}.
	\label{eq:region_channel_weights}
\end{equation}
The resulting modulation map is applied to the corresponding original matching features to obtain the adapted matching features $\hat{\boldsymbol{F}}_{i}^{M}$ as follows:
\begin{equation}
	\hat{\boldsymbol{F}}_{i}^{M}
	=\boldsymbol{F}_{i}^{M}
	+\lambda\,\boldsymbol{\Gamma}_{i}\odot\boldsymbol{F}_{i}^{M},
	\label{eq:region_channel_modulation}
\end{equation}
where $\lambda$ denotes a hyperparameter controlling the modulation strength.

\subsection{Spatial Reorganization via Attraction and Repulsion}
\label{3.4}

The analysis of local context feature affinity in Fig.~\ref{fig:motivation}(b) suggests that the original context features exhibit neither sufficient consistency within motion-coherent regions nor sufficient distinction across motion boundaries. Although adaptive aggregation of neighboring features~\cite{hu2019local,li2021involution} can improve local consistency, indiscriminately aggregating features across motion boundaries may blur these distinctions. This calls for spatial reorganization that promotes consistency among similar features while strengthening separation from dissimilar ones. Inspired by the attraction-and-repulsion mechanism proposed in the prior works~\cite{yu2001understanding,de2017semantic}, TO-Flow introduces two complementary convolution-based operations based on local affinity: one performs feature attraction by aggregating neighboring features with high affinity to promote local consistency, while the other performs feature repulsion by first aggregating neighboring features with low affinity to form a reference and then adding the difference between the current feature and this reference back to the current feature.

Specifically, two $3\times3$ convolutional layers, denoted by $\operatorname{Conv}^{B}(\cdot)$ and $\operatorname{Conv}^{I}(\cdot)$, first take $\boldsymbol{F}^{C}$ as input to produce kernel logit maps $\boldsymbol{\Phi}^{B}=\operatorname{Conv}^{B}(\boldsymbol{F}^{C})\in\mathbb{R}^{H^{C}\times W^{C}\times9}$ and $\boldsymbol{\Phi}^{I}=\operatorname{Conv}^{I}(\boldsymbol{F}^{C})\in\mathbb{R}^{H^{C}\times W^{C}\times25}$, respectively.
The affinity maps $\boldsymbol{A}_{3}$ and $\boldsymbol{A}_{5}$ obtained from $\boldsymbol{F}^{C}$ in Sect.~\ref{3.2} are then used to construct affinity priors $\boldsymbol{P}^{B}$ and $\boldsymbol{P}^{I}$ for boundary enhancement and interior propagation, respectively, as follows:
\begin{equation}
	\boldsymbol{P}^{B}(\boldsymbol{p},j)
	=\begin{cases}
		1,&j=0\\
		1-\boldsymbol{A}_{3}(\boldsymbol{p},j),&j=1,\ldots,8
	\end{cases},
	\label{eq:affinity_kernel_priors}
\end{equation}
and
\begin{equation}
	\boldsymbol{P}^{I}(\boldsymbol{p},j)
	=\begin{cases}
		1,&j=0\\
		\boldsymbol{A}_{5}(\boldsymbol{p},j),&j=1,\ldots,24
	\end{cases}.
	\label{eq:interior_kernel_prior}
\end{equation}
The affinity priors are explicitly incorporated into the generation of feature aggregation kernels $\boldsymbol{K}^{G}$ as follows:
\begin{equation}
	\begin{aligned}
		\boldsymbol{K}^{G}(\boldsymbol{p},j)
		=\frac{\boldsymbol{P}^{G}(\boldsymbol{p},j)
			\exp\!\left(\boldsymbol{\Phi}^{G}(\boldsymbol{p},j)\right)}
		{\displaystyle\sum_{j'}
			\boldsymbol{P}^{G}(\boldsymbol{p},j')
			\exp\!\left(\boldsymbol{\Phi}^{G}(\boldsymbol{p},j')\right)},
	\end{aligned}
	\label{eq:boundary_adaptive_kernel}
\end{equation}
where $\exp(\cdot)$ denotes the exponential function.
Within motion-coherent regions, $\boldsymbol{P}^{I}$ assigns higher weights to neighboring features within the aggregation kernel that are similar to the central feature, while $\boldsymbol{P}^{B}$ does the opposite.
Applying the kernels $\boldsymbol{K}^{G}$ to $\boldsymbol{F}^{C}$ yields the aggregated feature maps $\boldsymbol{V}^{G}$ as follows:
\begin{equation}
	\boldsymbol{V}^{G}(\boldsymbol{p})
	=\sum_{j=0}^{(k^{G})^{2}-1}
	\boldsymbol{K}^{G}(\boldsymbol{p},j)
	\boldsymbol{F}^{C}(\boldsymbol{p}+\boldsymbol{r}_{k^{G},j}),
	\label{eq:adaptive_context_aggregation}
\end{equation}
where $k^{B}=3$ and $k^{I}=5$ denote the neighborhood sizes for boundary enhancement and interior propagation, respectively.
Using the soft mask $\boldsymbol{M}$ obtained from $\boldsymbol{F}^{C}$ in Sect.~\ref{3.2}, with its weights broadcast across feature channels, the boundary enhancement operation $\operatorname{Boundary}(\cdot)$ and the interior propagation operation $\operatorname{Interior}(\cdot)$ produce feature maps $\boldsymbol{F}^{B}$ and $\boldsymbol{F}^{I}$, respectively, as follows:
\begin{equation}
	\begin{aligned}
		\boldsymbol{F}^{B}
		&=\operatorname{Boundary}(\boldsymbol{F}^{C})\\
		&=\boldsymbol{F}^{C}+\left(1-\boldsymbol{M}\right)\odot
		\left(\boldsymbol{F}^{C}-\boldsymbol{V}^{B}\right),
	\end{aligned}
	\label{eq:boundary_feature_contrast}
\end{equation}
and
\begin{equation}
	\begin{aligned}
		\boldsymbol{F}^{I}
		&=\operatorname{Interior}(\boldsymbol{F}^{C})\\
		&=\left(1-\boldsymbol{M}\right)\odot\boldsymbol{F}^{C}
		+\boldsymbol{M}\odot\boldsymbol{V}^{I}.
	\end{aligned}
	\label{eq:interior_feature_propagation}
\end{equation}

However, the two operations defined above may conflict when both are applied directly at the current resolution. While this relatively fine resolution preserves the boundary cues needed for feature repulsion, its limited receptive field restricts the aggregation of similar features and makes feature attraction more sensitive to local appearance variations. TO-Flow therefore applies and combines the two operations from coarse to fine.
To be specific, $\boldsymbol{F}^{C}$ is progressively downsampled to construct the feature pyramid $\{\boldsymbol{F}_{\ell}^{C}\}_{\ell=1}^{3}$, ordered from finest to coarsest, with $\boldsymbol{F}_{1}^{C}=\boldsymbol{F}^{C}$ and $\boldsymbol{F}_{\ell}^{C}\in\mathbb{R}^{\frac{H^{C}}{2^{\ell-1}}\times\frac{W^{C}}{2^{\ell-1}}\times D^{C}}$.
At each scale, the partitioning process in Sect.~\ref{3.2} generates the mask $\boldsymbol{M}_{\ell}$ and affinity maps $\boldsymbol{A}_{3,\ell}$ and $\boldsymbol{A}_{5,\ell}$ from $\boldsymbol{F}_{\ell}^{C}$ for $\operatorname{Boundary}_{\ell}(\cdot)$ and $\operatorname{Interior}_{\ell}(\cdot)$. For each pair of adjacent scales $\ell$ and $\ell-1$, with $\ell=3,2$, feature repulsion is applied directly to the finer features to increase the difference between features on opposite sides of motion boundaries as follows:
\begin{equation}
	\boldsymbol{F}_{\ell-1}^{B}
	=\operatorname{Boundary}_{\ell-1}\!\left(\boldsymbol{F}_{\ell-1}^{C}\right).
	\label{eq:finer_boundary_enhancement}
\end{equation}
Meanwhile, the coarser features are first processed by feature attraction to promote consistency within motion-coherent regions and then upsampled to the same resolution as the finer features, as follows:
\begin{equation}
	\boldsymbol{F}_{\ell-1}^{U}
	=\operatorname{Up}_{\ell-1}\!\left(
	\operatorname{Interior}_{\ell}(\boldsymbol{F}_{\ell}^{C})
	\right),
	\label{eq:region_feature_propagation}
\end{equation}
where $\operatorname{Up}_{\ell-1}(\cdot)$ denotes upsampling to the resolution of $\boldsymbol{F}_{\ell-1}^{C}$.

However, directly upsampling the coarser features can introduce boundary blurring, which may weaken the enhanced boundary contrast in the finer features when the two are fused. To address this issue, the upsampled coarser features are first resampled using local similarities and feature cues from the finer level before fusion.
Specifically, the refined similarity map $\widetilde{\boldsymbol{S}}_{3,\ell-1}$ obtained from $\boldsymbol{F}_{\ell-1}^{C}$ in Sect.~\ref{3.2} is normalized over the eight neighboring offsets at each pixel to obtain a weighting map $\boldsymbol{W}_{\ell-1}=\operatorname{Softmax}(\widetilde{\boldsymbol{S}}_{3,\ell-1})$.
This weighting map is used to compute a weighted average of the offset vectors from each pixel to its eight neighbors, yielding the sampling direction $\boldsymbol{d}_{\ell-1}(\boldsymbol{p})\in\mathbb{R}^{2}$ as follows:
\begin{equation}
	\boldsymbol{d}_{\ell-1}(\boldsymbol{p})
	=\sum_{j=1}^{8}\boldsymbol{W}_{\ell-1}(\boldsymbol{p},j)\boldsymbol{r}_{3,j}.
	\label{eq:resampling_direction}
\end{equation}
The sampling distance along this direction is predicted from the finer features and their refined similarities as follows:
\begin{equation}
	\boldsymbol{D}_{\ell-1}
	=\sigma\!\left(\operatorname{Conv}_{\ell-1}\!\left(
	\boldsymbol{F}_{\ell-1}^{C},\widetilde{\boldsymbol{S}}_{3,\ell-1}
	\right)\right),
	\label{eq:sampling_radius}
\end{equation}
where $\operatorname{Conv}_{\ell-1}(\cdot,\cdot)$ denotes a predictor consisting of two $3\times3$ convolutional layers.
The predicted distance and sampling direction determine the sampling location as follows:
\begin{equation}
	\boldsymbol{p}_{\ell-1}^{S}(\boldsymbol{p})
	=\boldsymbol{p}+\boldsymbol{D}_{\ell-1}(\boldsymbol{p})
	\frac{\boldsymbol{d}_{\ell-1}(\boldsymbol{p})}
	{\|\boldsymbol{d}_{\ell-1}(\boldsymbol{p})\|_2}.
	\label{eq:boundary_resampling}
\end{equation}
The mask $\boldsymbol{M}_{\ell-1}$ is then used to concentrate feature resampling near motion boundaries while largely preserving the original upsampled features within region interiors, as follows:
\begin{equation}
	\begin{aligned}
		\boldsymbol{F}_{\ell-1}^{S}(\boldsymbol{p})
		=&\left(1-\boldsymbol{M}_{\ell-1}(\boldsymbol{p})\right)\mathcal{W}\!\left(
		\boldsymbol{F}_{\ell-1}^{U},\boldsymbol{p}_{\ell-1}^{S}(\boldsymbol{p})\right) \\
		&+\boldsymbol{M}_{\ell-1}(\boldsymbol{p})\boldsymbol{F}_{\ell-1}^{U}(\boldsymbol{p})
	\end{aligned}.
	\label{eq:resampled_context_contribution}
\end{equation}
The resampled coarse features and the boundary-enhanced finer features are then fused as follows:
\begin{equation}
	\hat{\boldsymbol{F}}_{\ell-1}^{C}
	=\operatorname{Conv}_{\ell-1}^{F}\!\left(
	[\boldsymbol{F}_{\ell-1}^{B},\boldsymbol{F}_{\ell-1}^{S}]
	\right),
	\label{eq:multiscale_context_reorganization}
\end{equation}
where $[\cdot,\cdot]$ denotes concatenation along the channel dimension, and $\operatorname{Conv}_{\ell-1}^{F}(\cdot)$ projects the concatenated features to $D^{C}$ channels.
At each intermediate level, the fused feature map
$\hat{\boldsymbol{F}}_{\ell-1}^{C}$ serves as the coarser input to
\eqref{eq:region_feature_propagation} for the next pair of adjacent feature levels. Repeating the above procedure yields the final context features $\hat{\boldsymbol{F}}_{1}^{C}$ for recurrent flow refinement.

\section{Experiments}
\label{sec:experiments}

Sect.~\ref{impl} describes the implementation details of the TO-Flow framework across different architectures and the corresponding training settings. Sect.~\ref{data} introduces the datasets used in the subsequent experiments, while the evaluation metrics are described in Sect.~\ref{metr}. Sects.~\ref{sur} and~\ref{unsur} present comparisons of the TO-Flow variants with the corresponding baseline models and other representative methods in supervised and unsupervised optical flow estimation, respectively. Finally, Sect.~\ref{abl} presents ablation studies to assess the contributions of the key components and evaluate the design choices in TO-Flow.

\begin{table*}[t]
\centering
\settablefont
\setlength{\tabcolsep}{22.63pt}
\caption{Zero-shot generalization results on the MPI Sintel and KITTI 2015 training sets. Methods are grouped by training data. C, T, MD, YV, AF, and Tartan denote the FlyingChairs, FlyingThings3D, MegaDepth, YouTube-VOS, AutoFlow, and TartanAir datasets, respectively. TO-Flow ($\cdot$) denotes the corresponding baseline integrated with the TO-Flow framework. Better results between each baseline and its corresponding TO-Flow variant are marked in bold.}
\label{tb.zeroshot}
\begin{tabular}{@{}llccccc@{}}
			\toprule
			\multirow{2}{*}[-0.75ex]{Training Data} & \multirow{2}{*}[-0.75ex]{Method} & \multicolumn{2}{c}{Sintel (train)} & \multicolumn{2}{c}{KITTI 2015 (train)} & \multirow{2}{*}[-0.75ex]{Params (M)} \\
			\cmidrule(lr){3-4}\cmidrule(lr){5-6}
			& & Clean$\downarrow$ & Final$\downarrow$ & EPE$\downarrow$ & Fl-all$\downarrow$ & \\
			\midrule
			\multirow{3}{*}{Tartan+C+T} & GMFlow \settablecitationfont\clgray{CVPR'22} \cite{xu2022gmflow} & -- & -- & 8.70 & 24.40 & 4.7 \\
			& SEA-RAFT-M \settablecitationfont\clgray{ECCV'24} \cite{wang2024sea} & 1.27 & 3.85 & 4.30 & 14.30 & 19.7 \\
			& FlowSeek-T \settablecitationfont\clgray{ICCV'25} \cite{poggi2025flowseek} & 1.13 & 2.48 & 4.06 & 12.20 & 42.7 \\
			\midrule
			C+T+MD & MatchFlow-G \settablecitationfont\clgray{CVPR'23} \cite{dong2023rethinking} & 1.03 & 2.45 & 4.08 & 15.60 & 15.4 \\
			\midrule
			C+T+YV & FlowFormer++ \settablecitationfont\clgray{CVPR'23} \cite{shi2023flowformer++} & 0.90 & 2.30 & 3.93 & 14.13 & 16.1 \\
			\midrule
			AF+T & FlowDiffuser \settablecitationfont\clgray{CVPR'24} \cite{luo2024flowdiffuser} & 0.86 & 2.19 & 3.61 & 11.80 & -- \\
			\midrule
			\multirow{21}{*}{C+T}
			& PWC-Net \settablecitationfont\clgray{CVPR'18} \cite{sun2018pwc} & 2.55 & 3.93 & 10.40 & 33.70 & 9.4 \\
			& RAFT \settablecitationfont\clgray{ECCV'20} \cite{teed2020raft} & 1.43 & 2.71 & 5.04 & 17.40 & 5.3 \\
			& GMA \settablecitationfont\clgray{ICCV'21} \cite{jiang2021learning} & 1.30 & 2.74 & 4.69 & 17.10 & 5.9 \\
			& SKFlow \settablecitationfont\clgray{NeurIPS'22} \cite{sun2022skflow} & 1.22 & 2.46 & 4.27 & 15.50 & 6.3 \\
			& DIP \settablecitationfont\clgray{CVPR'22} \cite{zheng2022dip} & 1.30 & 2.82 & 4.29 & 13.70 & 5.4 \\
			& EMD-L \settablecitationfont\clgray{ICCV'23} \cite{deng2023explicit} & 0.88 & 2.55 & 4.12 & 13.50 & 13.7 \\
			& CRAFT \settablecitationfont\clgray{CVPR'22} \cite{sui2022craft} & 1.27 & 2.79 & 4.88 & 17.50 & 6.3 \\
			& RPKNet \settablecitationfont\clgray{AAAI'24} \cite{morimitsu2024recurrent} & 1.12 & 2.45 & -- & 13.00 & 2.8 \\
			& GMFlowNet \settablecitationfont\clgray{CVPR'22} \cite{zhao2022global} & 1.14 & 2.71 & 4.24 & 15.40 & 9.3 \\
			& FlowFormer \settablecitationfont\clgray{ECCV'22} \cite{huang2022flowformer} & 1.01 & 2.40 & 4.09 & 14.70 & 16.2 \\
			& CCMR+ \settablecitationfont\clgray{WACV'24} \cite{jahedi2024ccmr} & 0.98 & 2.36 & -- & 12.90 & 11.5 \\
			& AnyFlow \settablecitationfont\clgray{CVPR'23} \cite{jung2023anyflow} & 1.10 & 2.52 & 3.76 & 12.40 & 5.4 \\
			& SAMFlow \settablecitationfont\clgray{AAAI'24} \cite{zhou2024samflow} & 0.87 & 2.11 & 3.44 & 12.30 & -- \\
			& SEA-RAFT-M \settablecitationfont\clgray{ECCV'24} \cite{wang2024sea} & 1.21 & 4.04 & 4.29 & 14.20 & 19.7 \\
			& DPFlow \settablecitationfont\clgray{CVPR'25} \cite{morimitsu2025dpflow} & 1.02 & 2.26 & 3.37 & 11.10 & 10.0 \\
			& GMFlow+ \settablecitationfont\clgray{TPAMI'23} \cite{xu2023unifying} & 0.91 & 2.74 & 5.74 & 17.60 & 7.6 \\
			& OFM \settablecitationfont\clgray{CVPR'26} \cite{luo2026optical} & 0.81 & 2.16 & 3.32 & 10.90 & 15.6 \\
			\cdashline{2-7}
			& FlowSeek-T \settablecitationfont\clgray{ICCV'25} \cite{poggi2025flowseek} & \textbf{1.12} & 2.53 & 3.95 & 12.70 & 42.7 \\
			& \textbf{TO-Flow (FlowSeek-T)} & 1.17 & \textbf{2.43} & \textbf{3.87} & \textbf{12.20} & 46.9 \\
			\cdashline{2-7}
			& WAFT-DAv2-a2 \settablecitationfont\clgray{ICLR'26} \cite{wang2026waft} & 1.01 & 2.49 & 3.28 & 10.90 & 56.9 \\
			& \textbf{TO-Flow (WAFT-DAv2-a2)} & \textbf{0.98} & \textbf{2.30} & \textbf{2.96} & \textbf{10.20} & 64.7 \\
			\bottomrule
\end{tabular}
\end{table*}

\begin{table*}[t]
\centering
\settablefont
\setlength{\tabcolsep}{16.92pt}
\caption{Fine-tuned results on the MPI Sintel and KITTI 2015 training and test sets. Better results between each baseline and its corresponding TO-Flow variant are marked in bold. Ties at the reported precision are also marked in bold.}
\label{tb.fine-tune}
\begin{tabular}{@{}lccccccc@{}}
			\toprule
			\multirow{2}{*}[-0.75ex]{Method} & \multicolumn{2}{c}{Sintel (train)} & \multicolumn{2}{c}{KITTI 2015 (train)} & \multicolumn{2}{c}{Sintel (test)} & \multicolumn{1}{c}{KITTI 2015 (test)} \\
			\cmidrule(lr){2-3}\cmidrule(lr){4-5}\cmidrule(lr){6-7}\cmidrule(lr){8-8}
			& Clean$\downarrow$ & Final$\downarrow$ &  EPE$\downarrow$ & Fl-all$\downarrow$ & Clean$\downarrow$ & Final$\downarrow$ & Fl-all$\downarrow$ \\
			\midrule
			PWC-Net \settablecitationfont\clgray{CVPR'18} \cite{sun2018pwc} & 1.70 & 2.21 & 2.16 & 9.80 & 3.86 & 5.13 & 9.60 \\
			RAFT \settablecitationfont\clgray{ECCV'20} \cite{teed2020raft} & 0.76 & 1.22 & 0.63 & 1.50 & 1.61 & 2.86 & 5.10 \\
			GMA \settablecitationfont\clgray{ICCV'21} \cite{jiang2021learning} & 0.62 & 1.06 & 0.57 & 1.20 & 1.39 & 2.47 & 5.15 \\
			SKFlow \settablecitationfont\clgray{NeurIPS'22} \cite{sun2022skflow} & 0.52 & 0.78 & 0.51 & 0.94 & 1.28 & 2.23 & 4.84 \\
			DIP \settablecitationfont\clgray{CVPR'22} \cite{zheng2022dip} & -- & -- & -- & -- & 1.44 & 2.83 & 4.21 \\
			EMD-L \settablecitationfont\clgray{ICCV'23} \cite{deng2023explicit} & -- & -- & -- & -- & 1.32 & 2.51 & 4.49 \\
			CRAFT \settablecitationfont\clgray{CVPR'22} \cite{sui2022craft} & 0.60 & 1.06 & 0.58 & 1.34 & 1.45 & 2.42 & 4.79 \\
			RPKNet \settablecitationfont\clgray{AAAI'24} \cite{morimitsu2024recurrent} & -- & -- & -- & -- & 1.31 & 2.65 & 4.64 \\
			GMFlowNet \settablecitationfont\clgray{CVPR'22} \cite{zhao2022global} & 0.59 & 0.91 & 0.64 & 1.51 & 1.39 & 2.65 & 4.79 \\
			FlowFormer \settablecitationfont\clgray{ECCV'22} \cite{huang2022flowformer} & 0.48 & 0.74 & 0.53 & 1.11 & 1.20 & 2.12 & 4.68 \\
			FlowFormer++ \settablecitationfont\clgray{CVPR'23} \cite{shi2023flowformer++} & 0.40 & 0.60 & 0.57 & 1.16 & 1.07 & 1.94 & 4.52 \\
			CCMR+ \settablecitationfont\clgray{WACV'24} \cite{jahedi2024ccmr} & -- & -- & -- & -- & 1.07 & 2.10 & 3.86 \\
			MatchFlow-G \settablecitationfont\clgray{CVPR'23} \cite{dong2023rethinking} & 0.49 & 0.78 & 0.55 & 1.10 & 1.16 & 2.37 & 4.63 \\
			AnyFlow \settablecitationfont\clgray{CVPR'23} \cite{jung2023anyflow} & -- & -- & -- & -- & 1.23 & 2.44 & 4.41 \\
			SAMFlow \settablecitationfont\clgray{AAAI'24} \cite{zhou2024samflow} & -- & -- & -- & -- & 1.00 & 2.08 & 4.49 \\
			FlowDiffuser \settablecitationfont\clgray{CVPR'24} \cite{luo2024flowdiffuser} & -- & -- & -- & -- & 1.02 & 2.03 & 4.17 \\
			DPFlow \settablecitationfont\clgray{CVPR'25} \cite{morimitsu2025dpflow} & -- & -- & -- & -- & 1.04 & 1.97 & 3.56 \\
			SEA-RAFT-M  \settablecitationfont\clgray{ECCV'24} \cite{wang2024sea} & -- & -- & -- & -- & 1.44 & 2.86 & 4.64 \\
			GMFlow+ \settablecitationfont\clgray{TPAMI'23} \cite{xu2023unifying} & -- & -- & -- & -- & 1.03 & 2.37 & 4.49 \\
			MS-RAFT+ \settablecitationfont\clgray{IJCV'24} \cite{jahedi2024ms} & -- & -- & -- & -- & 1.23 & 2.68 & 4.15 \\
			OFM \settablecitationfont\clgray{CVPR'26} \cite{luo2026optical} & -- & -- & -- & -- & 0.94 & 1.85 & 3.78 \\
			\hdashline
			FlowSeek-T \settablecitationfont\clgray{ICCV'25} \cite{poggi2025flowseek} & 0.57 & 0.92 & 1.02 & 3.02 & 1.64 & \textbf{3.00} & 5.38 \\
			\textbf{TO-Flow (FlowSeek-T)} & \textbf{0.51} & \textbf{0.82} & \textbf{0.98} & \textbf{2.97} & \textbf{1.43} & 3.04 & \textbf{4.66} \\
			\hdashline
			WAFT-DAv2-a2 \settablecitationfont\clgray{ICLR'26} \cite{wang2026waft} & \textbf{0.38} & \textbf{0.56} & 0.65 & 1.51 & 0.95 & \textbf{2.33} & 3.31 \\
			\textbf{TO-Flow (WAFT-DAv2-a2)} & \textbf{0.38} & 0.58 & \textbf{0.64} & \textbf{1.47} & \textbf{0.90} & 2.35 & \textbf{3.22} \\
			\bottomrule
\end{tabular}
\end{table*}

\begin{figure*}[!t]
	\centering
	\includegraphics[width=0.99\textwidth]{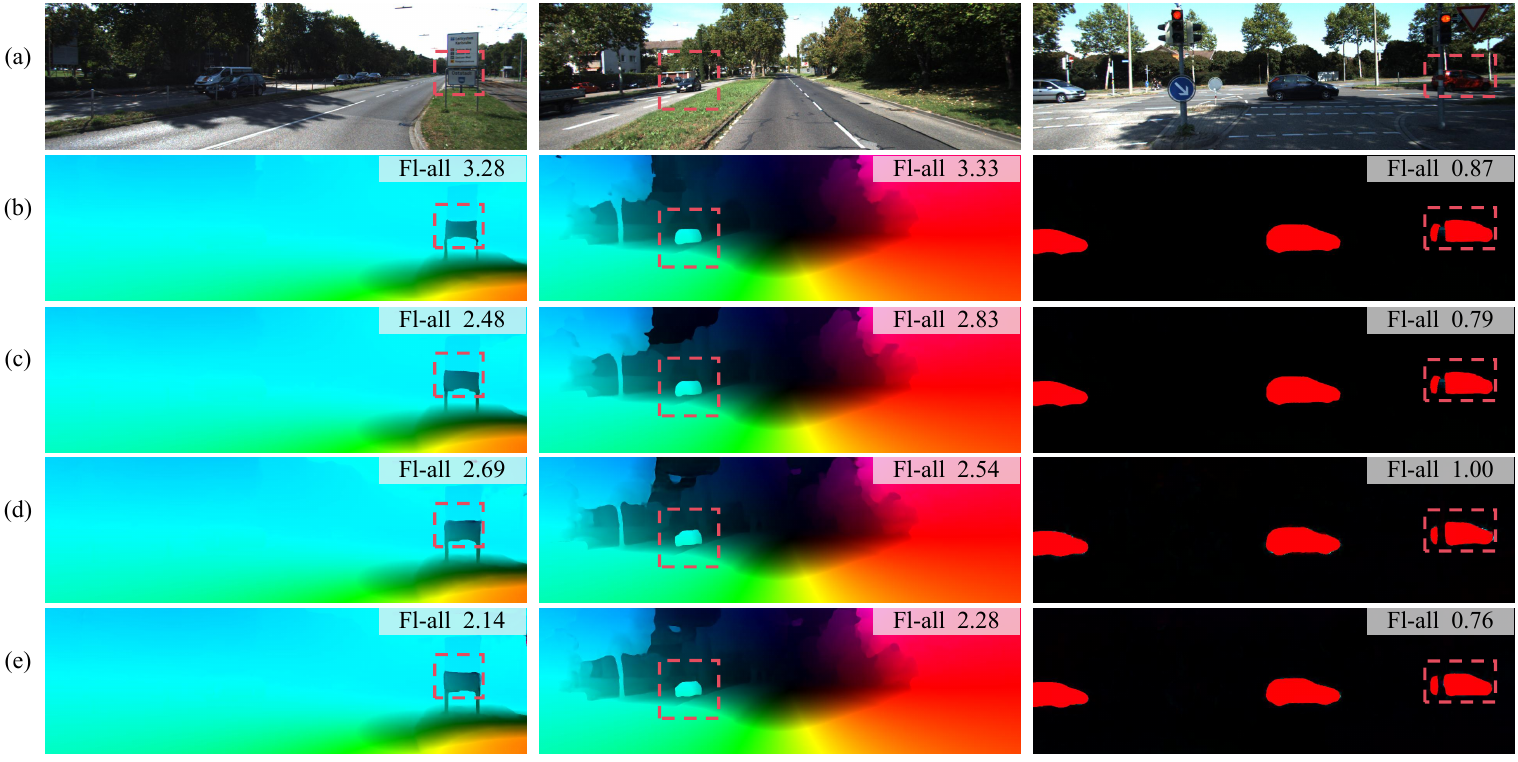}
	\caption{
		Qualitative comparison on the KITTI 2015 dataset. Rows show (a) the first frame, (b) FlowSeek-T, (c) TO-Flow (FlowSeek-T), (d) WAFT-DAv2-a2, and (e) TO-Flow (WAFT-DAv2-a2). Dashed boxes indicate corresponding regions of interest. Fl-all values are reported for each example; lower values indicate better performance.
	}
	\label{kitti}
\end{figure*}

\begin{figure*}[t]
	\centering
	\includegraphics[width=0.99\textwidth]{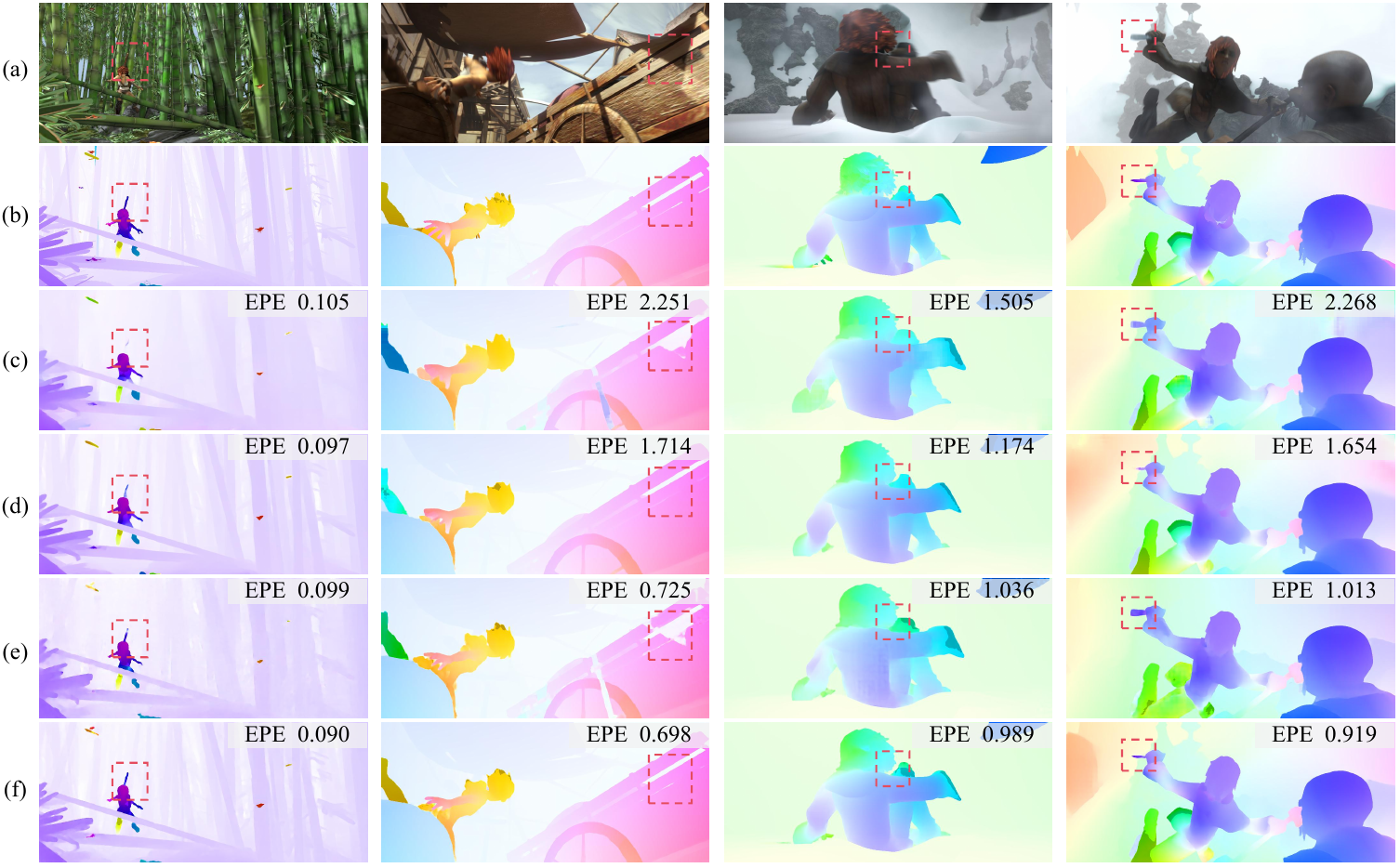}
	\caption{
		Qualitative comparison on the MPI Sintel dataset. Rows show (a) the first frame, (b) ground-truth flow, (c) FlowSeek-T, (d) TO-Flow (FlowSeek-T), (e) WAFT-DAv2-a2, and (f) TO-Flow (WAFT-DAv2-a2). Dashed boxes indicate corresponding regions of interest. EPE values are reported for each example; lower values indicate better performance.
	}
	\label{sintel}
\end{figure*}

\begin{figure*}[t]
	\centering
	\includegraphics[width=0.99\textwidth]{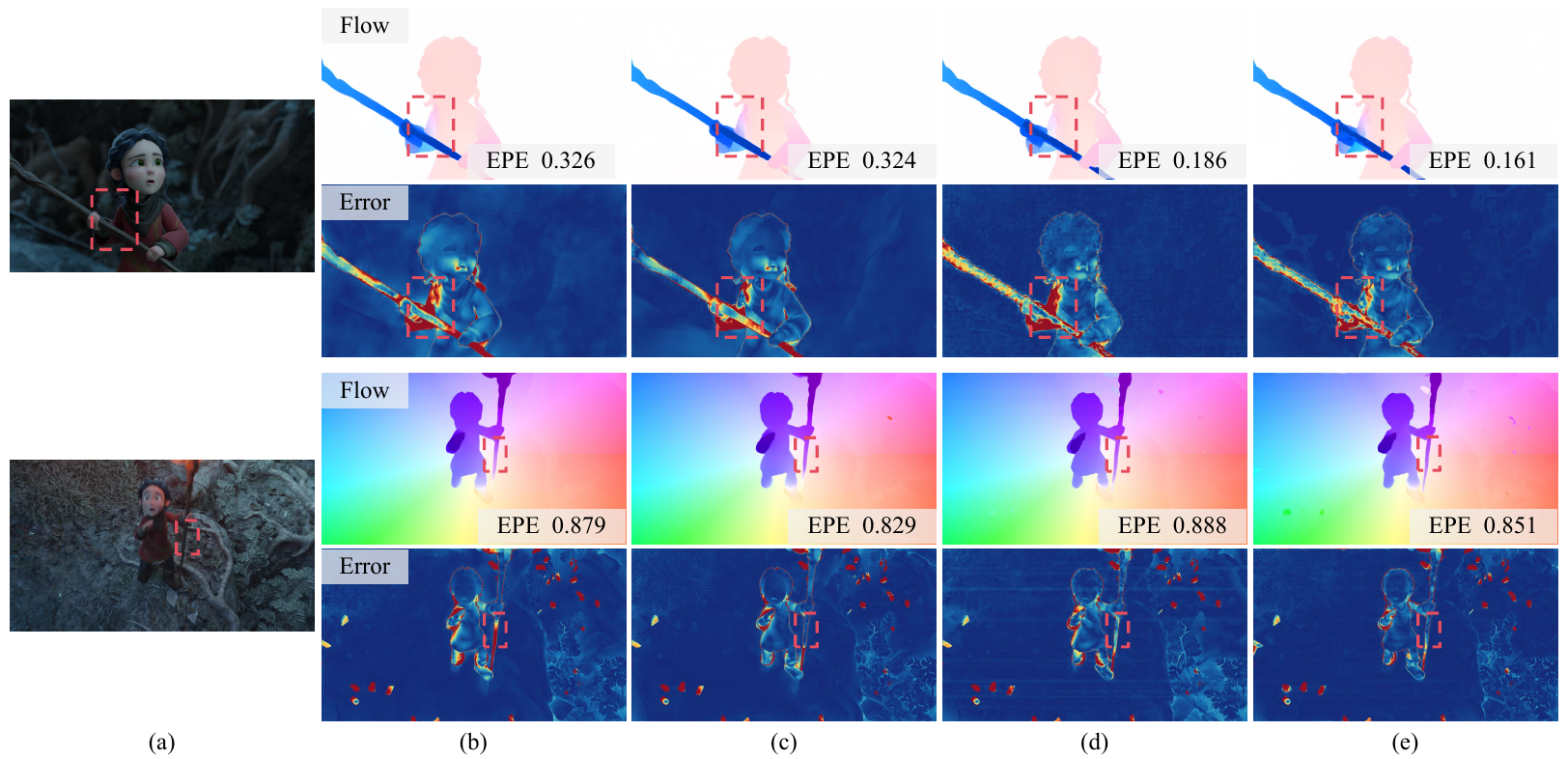}
	\caption{Qualitative comparison on the Spring dataset. Columns show (a) the first frame, (b) FlowSeek-T, (c) TO-Flow (FlowSeek-T), (d) WAFT-DAv2-a2, and (e) TO-Flow (WAFT-DAv2-a2). For each example, predicted flow maps and corresponding error maps are shown in the upper and lower rows, respectively. Dashed boxes highlight fine structures. EPE values are reported for each example; lower values indicate better performance.}
	\label{spring}
\end{figure*}

\begin{figure*}[!t]
	\centering
	\includegraphics[width=0.99\textwidth]{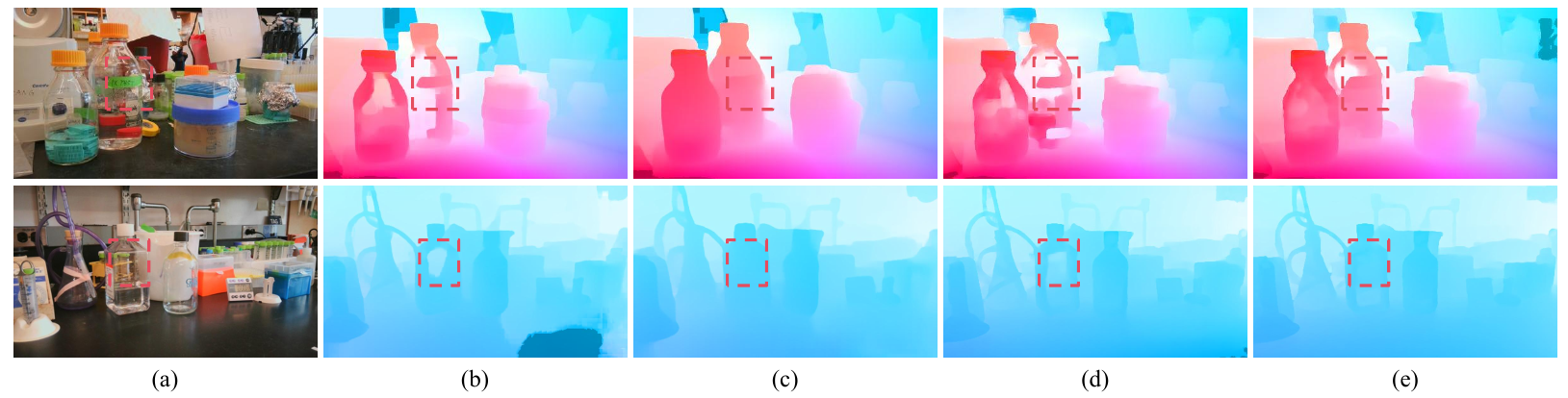}
	\caption{
		Qualitative comparison on the LayeredFlow dataset. Columns show (a) the first frame, (b) FlowSeek-T, (c) TO-Flow (FlowSeek-T), (d) WAFT-DAv2-a2, and (e) TO-Flow (WAFT-DAv2-a2). Dashed boxes highlight corresponding regions containing transparent objects.
	}
	\label{layeredflow}
\end{figure*}

\begin{table*}[t]
\centering
\settablefont
\setlength{\tabcolsep}{18.05pt}
\caption{Zero-shot generalization results on the LayeredFlow validation set and Spring training set. Methods are grouped by training data. VIPER denotes the VIsual PERception benchmark, and FA-Flow denotes the dataset introduced in Flow-Anything. Better results between each baseline and its corresponding TO-Flow variant are marked in bold.}
\label{tb.layered_spring}
\begin{tabular}{@{}llcccccc@{}}
			\toprule
			\multirow{2}{*}[-0.75ex]{Training Data} & \multirow{2}{*}[-0.75ex]{Method} & \multicolumn{4}{c}{LayeredFlow (val)} & \multicolumn{2}{c}{Spring (train)} \\
			\cmidrule(lr){3-6}\cmidrule(lr){7-8}
			& & EPE$\downarrow$ & 1px$\downarrow$ & 3px$\downarrow$ & 5px$\downarrow$ & 1px$\downarrow$ & EPE$\downarrow$ \\
			\midrule
			C+T+S & PWC-Net \settablecitationfont\clgray{CVPR'18} \cite{sun2018pwc} & 9.49 & 74.93 & 50.47 & 39.05 & -- & -- \\
			\midrule
			MD+C+T+TSKH & MatchFlow-G \settablecitationfont\clgray{CVPR'23} \cite{dong2023rethinking} & -- & -- & -- & -- & 4.504 & 0.407 \\
			\midrule
			YV+C+T+TSKH & FlowFormer++ \settablecitationfont\clgray{CVPR'23} \cite{shi2023flowformer++} & -- & -- & -- & -- & 4.482 & 0.447 \\
			\midrule
			C+T+TSKH+VIPER & MS-RAFT+ \settablecitationfont\clgray{IJCV'24} \cite{jahedi2024ms} & -- & -- & -- & -- & 3.577 & 0.397 \\
			\midrule
			C+T+S+K+H+FA-Flow & Flow-Anything \settablecitationfont\clgray{TPAMI'25} \cite{liang2025flow} & -- & -- & -- & -- & 3.790 & 0.398 \\
			\midrule
			\multirow{15}{*}{C+T+TSKH}
			& SEA-RAFT-M  \settablecitationfont\clgray{ECCV'24} \cite{wang2024sea} & 10.17 & 69.73 & 45.94 & 34.78 & 3.888 & 0.406 \\
			& ReCoVEr-CX  \settablecitationfont\clgray{ICCV'25} \cite{kiefhaber2025removing} & -- & -- & -- & -- & -- & 0.500 \\
			& RAFT \settablecitationfont\clgray{ECCV'20} \cite{teed2020raft} & 9.38 & 71.98 & 46.46 & 36.15 & 4.788 & 0.448 \\
			& GMA \settablecitationfont\clgray{ICCV'21} \cite{jiang2021learning} & 9.77 & 72.46 & 46.93 & 36.97 & 4.763 & 0.443 \\
			& SKFlow \settablecitationfont\clgray{NeurIPS'22} \cite{sun2022skflow} & 9.86 & 72.02 & 47.44 & 36.88 & 4.521 & 0.408 \\
			& DIP \settablecitationfont\clgray{CVPR'22} \cite{zheng2022dip} & -- & -- & -- & -- & 4.273 & 0.463 \\
			& CRAFT \settablecitationfont\clgray{CVPR'22} \cite{sui2022craft} & 10.36 & 72.34 & 47.54 & 37.00 & 4.803 & 0.448 \\
			& RPKNet \settablecitationfont\clgray{AAAI'24} \cite{morimitsu2024recurrent} & -- & -- & -- & -- & 4.472 & 0.416 \\
			& FlowFormer \settablecitationfont\clgray{ECCV'22} \cite{huang2022flowformer} & 10.20 & 73.59 & 48.97 & 38.56 & 4.508 & 0.470 \\
			& GMFlow+ \settablecitationfont\clgray{TPAMI'23} \cite{xu2023unifying} & 9.46 & 82.71 & 53.14 & 39.70 & 4.292 & 0.433 \\
			\cdashline{2-8}
			& FlowSeek-T \settablecitationfont\clgray{ICCV'25} \cite{poggi2025flowseek} & 9.09 & 70.82 & 43.74 & 32.36 & 4.111 & 0.410 \\
			& \textbf{TO-Flow (FlowSeek-T)} & \textbf{8.98} & \textbf{69.45} & \textbf{42.58} & \textbf{32.20} & \textbf{4.095} & \textbf{0.402} \\
			\cdashline{2-8}
			& WAFT-DAv2-a2 \settablecitationfont\clgray{ICLR'26} \cite{wang2026waft} & 11.40 & 72.04 & \textbf{47.92} & \textbf{37.76} & 3.358 & 0.360 \\
			& \textbf{TO-Flow (WAFT-DAv2-a2)} & \textbf{10.44} & \textbf{71.97} & 48.26 & 37.96 & \textbf{3.328} & \textbf{0.358} \\
			\bottomrule
\end{tabular}
\end{table*}

\subsection{Implementation Details}
\label{impl}

As TO-Flow is designed as a general framework for front-end feature adaptation, it is integrated into FlowSeek-T and WAFT-DAv2-a2 for supervised learning and U$^{2}$Flow without bidirectional flow fusion for unsupervised learning. Among these models, WAFT-DAv2-a2 is also used as the baseline for the ablation studies.

To ensure a fair comparison in both supervised and unsupervised experiments, the training settings for each TO-Flow variant follow those adopted in the corresponding baseline, including the optimizer, learning rate schedule, batch size, input resolution, and number of training iterations. For the ablation studies, the batch size is reduced to 4 to limit GPU memory usage, and the peak learning rate is set to $5 \times 10^{-5}$. The remaining optimization settings follow those of WAFT-DAv2-a2.

In terms of memory consumption, the TO-Flow framework introduces limited additional overhead. Taking FlowSeek-T as an example, TO-Flow increases peak training memory from 18.11~GiB to 21.01~GiB, an increase of approximately 16.0\%, with an input resolution of $432 \times 960$ pixels and a batch size of 4 in FP32. Both models can be trained on a single RTX 4090 GPU under these settings.

The feature-smoothing threshold $\theta$ and boundary threshold $\theta_{\mathrm{b}}$ are set to $0.2$ and $0.7$, respectively. For channel modulation of matching features, the number of simulated refinement iterations $T$ is set to $4$ for FlowSeek-T and $5$ for WAFT-DAv2-a2, matching the number of main flow refinement iterations used by the respective baselines. For U$^{2}$Flow, $T$ is set to $4$ rather than the full 12 refinement iterations to reduce the computational overhead of feedback collection. The residual scaling factor $\lambda$ is set to $0.02$ to control the strength of channel modulation. Spatial reorganization of context features is performed at three scales.

\subsection{Datasets}
\label{data}

To evaluate TO-Flow under supervised and unsupervised training paradigms and conduct ablation studies, nine public datasets are used for training and evaluation. These datasets are summarized as follows:

\begin{itemize}
	\item The \textbf{FlyingChairs}~\cite{dosovitskiy2015flownet} dataset contains 22,872 synthetic image pairs generated by applying transformations to chair images placed on natural backgrounds.

	\item The \textbf{FlyingThings3D}~\cite{mayer2016large} dataset provides 2,250 synthetic training sequences featuring diverse 3D object and camera motions.

	\item The \textbf{MPI Sintel}~\cite{butler2012naturalistic} dataset is a synthetic optical flow benchmark derived from an animated movie, comprising 23 training sequences and 12 test sequences.

	\item The \textbf{KITTI 2012 and KITTI 2015} datasets~\cite{geiger2012we,menze2015object} are real-world optical flow benchmarks covering outdoor driving scenes. Their training sets contain 194 and 200 image pairs, respectively, while their test sets contain 195 and 200 image pairs, respectively.

	\item The \textbf{KITTI Raw}~\cite{geiger2013vision} dataset provides image sequences captured from a moving vehicle in real-world driving scenes.

	\item The \textbf{HD1K}~\cite{kondermann2016hci} dataset contains 1,000 training frames from high-resolution real-world driving sequences. Following prior work~\cite{teed2020raft}, its training set (H) is combined with those of FlyingThings3D (T), MPI Sintel (S), and KITTI 2015 (K) to form the TSKH training set.

	\item The \textbf{Spring}~\cite{mehl2023spring} dataset is a high-resolution synthetic benchmark containing detailed scene structures and complex motions. Its training and test sets contain 37 and 10 sequences, respectively.

	\item The \textbf{LayeredFlow}~\cite{wen2024layeredflow} dataset is a real-world benchmark covering 185 indoor and outdoor scenes, featuring non-Lambertian objects with transparent or reflective surfaces.
\end{itemize}

For the supervised experiments, each TO-Flow variant follows the training schedule adopted for its corresponding baseline. TO-Flow (FlowSeek-T) is trained for 100k iterations on the FlyingChairs dataset and then for 120k iterations on the FlyingThings3D dataset. The corresponding training stages for TO-Flow (WAFT-DAv2-a2) last 50k and 100k iterations, respectively. Generalization to unseen datasets is then evaluated on the training sets of the MPI Sintel and KITTI 2015 datasets.
The two variants are subsequently fine-tuned on TSKH for 300k and 200k iterations, respectively. For evaluation on the KITTI 2015 dataset, TO-Flow (WAFT-DAv2-a2) is instead fine-tuned only on its training set for 5k iterations. Performance is then evaluated on both the training and test sets of the MPI Sintel and KITTI 2015 datasets. To assess generalization beyond these commonly used benchmarks, the two variants fine-tuned on TSKH are further evaluated on the LayeredFlow validation set and the Spring training set. On the LayeredFlow dataset, evaluation is restricted to annotated points on the first surface intersected by each corresponding camera ray.

For the unsupervised experiments, the training and evaluation strategies follow those adopted in U$^{2}$Flow \cite{sun2026u}. For experiments on the MPI Sintel dataset, TO-Flow (U$^{2}$Flow) is first trained for 100k iterations using 14,570 image pairs constructed from raw Sintel movie frames \cite{liu2020learning} and then fine-tuned for 50k iterations on the training set of the MPI Sintel dataset. The same training set is used for evaluation. For experiments on the KITTI 2015 dataset, the model is first trained for 100k iterations on the KITTI Raw dataset and then fine-tuned for 100k iterations using the multi-view extensions of the KITTI 2012 and KITTI 2015 datasets \cite{geiger2012we,menze2015object}, which provide additional frames before and after the original benchmark image pairs. Performance is evaluated on the training set of the KITTI 2015 dataset.

For the ablation studies, all models are trained for 50k iterations on the FlyingChairs dataset followed by 50k iterations on the FlyingThings3D dataset. The trained models are then evaluated on the MPI Sintel and KITTI 2015 training sets.

\subsection{Evaluation Metrics}
\label{metr}
The average end-point error (EPE), defined as the mean Euclidean distance between predicted and ground-truth flow vectors, is used to evaluate optical flow accuracy. Additionally, the percentages of pixels with end-point errors exceeding 1 pixel on the Spring dataset and 1, 3, and 5 pixels on the LayeredFlow dataset are reported. For the KITTI 2015 dataset, Fl-all is reported as the percentage of valid pixels whose end-point errors exceed both 3 pixels and 5\% of the ground-truth flow magnitude.

\begin{figure*}[!t]
	\centering
	\includegraphics[width=\textwidth]{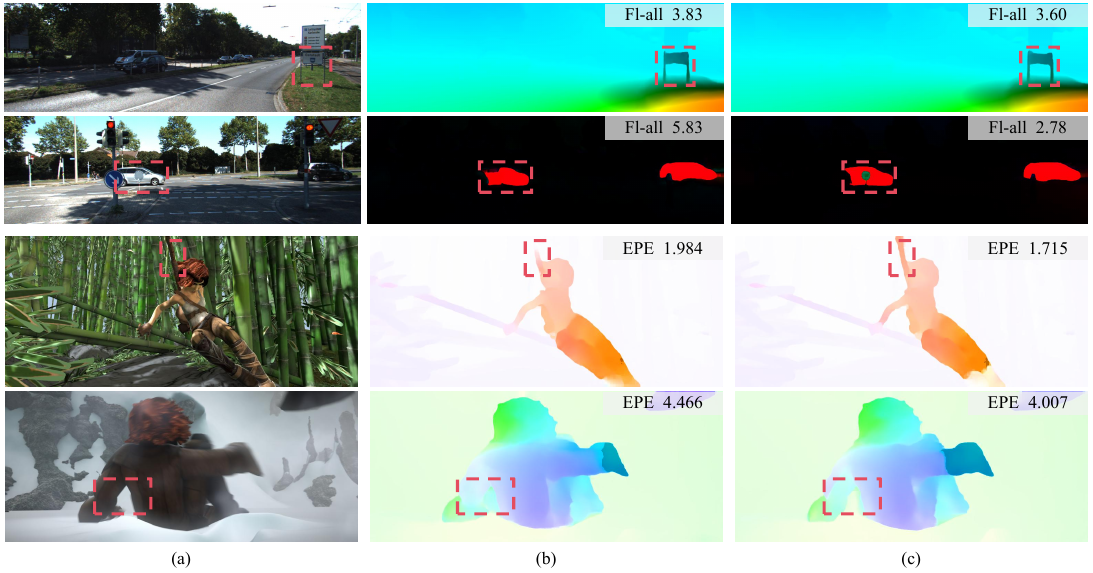}%
	\caption{Qualitative comparison between the reproduced U$^{2}$Flow baseline and TO-Flow (U$^{2}$Flow) on the KITTI 2015 (top two rows) and MPI Sintel (bottom two rows) datasets. Columns show (a) the first frame, (b) the baseline, and (c) TO-Flow. Dashed boxes highlight corresponding regions. Fl-all and EPE are reported for the KITTI 2015 and MPI Sintel examples, respectively; lower values indicate better performance.}
	\label{unsupvised}
\end{figure*}

\subsection{Supervised Optical Flow Estimation Results}
\label{sur}

This section evaluates TO-Flow in a supervised learning setting. TO-Flow is integrated into FlowSeek-T and WAFT-DAv2-a2, two recent optical flow models with competitive performance and distinct iterative flow refinement mechanisms, to assess its effectiveness and applicability across different architectures. The results obtained by the two TO-Flow variants are mainly compared with those of their respective baseline models. Results from other representative optical flow methods are also reported for reference.

Zero-shot generalization is first evaluated on the MPI Sintel and KITTI 2015 training sets. As shown in Table~\ref{tb.zeroshot}, the TO-Flow framework improves performance on most reported metrics for both baseline models. In particular, applying TO-Flow to WAFT-DAv2-a2 reduces the EPE on the Final pass of the MPI Sintel training set from 2.49 to 2.30 pixels and the Fl-all error rate on the KITTI 2015 training set from 10.90\% to 10.20\%, while increasing the total parameter count from 56.9M to 64.7M. These improvements suggest that the feature representations adapted by TO-Flow are better suited for optical flow estimation and improve cross-dataset generalization for both baseline models.
The models are subsequently fine-tuned following the training schedules of their respective baselines, and their flow predictions for the test sets are submitted to the official evaluation servers. As presented in Table~\ref{tb.fine-tune}, TO-Flow improves performance on most reported metrics, reducing the Fl-all error rate on the KITTI 2015 test set from 5.38\% to 4.66\% for FlowSeek-T and from 3.31\% to 3.22\% for WAFT-DAv2-a2. Nonetheless, in the zero-shot evaluation reported in Table~\ref{tb.zeroshot}, TO-Flow does not consistently improve performance on the Clean pass of MPI Sintel. This may be because the proposed feature adaptation mainly addresses ambiguities in matching and context features that become more pronounced under challenging conditions, while the clearer appearance cues in the Clean pass leave less room for improvement through feature adaptation.

Qualitative comparisons on the KITTI 2015 and MPI Sintel datasets are presented in Figs.~\ref{kitti} and~\ref{sintel}. Compared with the corresponding baselines, the optical flow maps predicted by both TO-Flow variants are more consistent in the interiors of motion-coherent regions and exhibit clearer motion boundaries, as highlighted by the red dashed boxes. These observations suggest that the proposed spatial reorganization provides more reliable motion cues for flow refinement by aggregating similar features in region interiors and enhancing their contrast with dissimilar neighbors near boundaries.

The Spring dataset contains fine structures and large displacements, while transparent and reflective surfaces in the LayeredFlow dataset can make visual correspondences ambiguous. These datasets are therefore used to assess whether TO-Flow can improve optical flow accuracy under such challenging conditions. As shown in Table~\ref{tb.layered_spring}, TO-Flow reduces both EPE and the 1-pixel error rate for both baseline models on the Spring training set. On the LayeredFlow validation set, the EPE decreases from 9.09 to 8.98 pixels for FlowSeek-T and from 11.40 to 10.44 pixels for WAFT-DAv2-a2, although the latter shows slight increases in the 3- and 5-pixel error rates. Overall, these results suggest that the TO-Flow framework remains beneficial when fine spatial details or ambiguous appearance cues complicate correspondence estimation.

On the Spring training set, the flow and error maps in Fig.~\ref{spring} show more accurate flow estimates for the highlighted fine structures. On the LayeredFlow validation set, the highlighted regions in Fig.~\ref{layeredflow} show that the predicted flow maps in the transparent bottle regions are more consistent with the bottle structure than with the visible background. These improvements may be explained by modulating feature channels according to their matching behavior, which helps emphasize discriminative matching cues even when fine structures or transparent surfaces make local appearance ambiguous.

\begin{table}[!t]
\centering
\settablefont
\setlength{\tabcolsep}{14.64pt}
\caption{Quantitative results on the MPI Sintel training set for unsupervised optical flow estimation. S-Raw denotes raw Sintel movie frames, and MF denotes multi-frame optical flow estimation. Better results between the reproduced baseline and its corresponding TO-Flow variant are marked in bold.}
\label{tb.unsupervised_sintel}
\begin{tabular}{@{}llcc@{}}
		\toprule
		Training Data & Method & Clean$\downarrow$ & Final$\downarrow$ \\
		\midrule
		C+S & UFlow \settablecitationfont\clgray{ECCV'20} \cite{jonschkowski2020matters}
		& 2.50 & 3.39 \\
		\midrule
		S & UPFlow \settablecitationfont\clgray{CVPR'21} \cite{luo2021upflow}
		& 2.33 & 2.67 \\
		\midrule
		\multirow{6}{*}{S-Raw+S}
		& SelFlow (MF) \settablecitationfont\clgray{CVPR'19} \cite{liu2019selflow}
		& 2.88 & 3.87 \\
		& ARFlow \settablecitationfont\clgray{CVPR'20} \cite{liu2020learning}
		& 2.79 & 3.73 \\
		& UnSAMFlow \settablecitationfont\clgray{CVPR'24} \cite{yuan2024unsamflow}
		& 2.21 & 3.07 \\
		\cdashline{2-4}
		& U$^{2}$Flow \settablecitationfont\clgray{CVPR'26} \cite{sun2026u}
		& 1.42 & 2.32 \\
		& U$^{2}$Flow (reproduced)
		& 1.59 & 2.51 \\
		& \textbf{TO-Flow (U$^{2}$Flow)}
		& \textbf{1.53} & \textbf{2.43} \\
		\bottomrule
\end{tabular}
\end{table}

\begin{table}[!t]
\centering
\settablefont
\setlength{\tabcolsep}{14.5pt}
\caption{Quantitative results on the KITTI 2015 training set for unsupervised optical flow estimation. K-Raw denotes the KITTI Raw dataset. K-MV denotes the multi-view extensions of the KITTI 2012 and KITTI 2015 datasets, while K15-MV denotes only the KITTI 2015 multi-view extension. Better results between the reproduced baseline and its corresponding TO-Flow variant are marked in bold.}
\label{tb.unsupervised_kitti}
\begin{tabular}{@{}llcc@{}}
		\toprule
		Training Data & Method & EPE$\downarrow$ & Fl-all$\downarrow$ \\
		\midrule
		K-MV & SelFlow (MF) \settablecitationfont\clgray{CVPR'19} \cite{liu2019selflow}
		& 4.84 & -- \\
		\midrule
		C+K15-MV & UFlow \settablecitationfont\clgray{ECCV'20} \cite{jonschkowski2020matters}
		& 2.71 & 9.05 \\
		\midrule
		\multirow{6}{*}{K-Raw+K-MV}
		& UPFlow \settablecitationfont\clgray{CVPR'21} \cite{luo2021upflow}
		& 2.45 & -- \\
		& ARFlow \settablecitationfont\clgray{CVPR'20} \cite{liu2020learning}
		& 2.85 & -- \\
		& UnSAMFlow \settablecitationfont\clgray{CVPR'24} \cite{yuan2024unsamflow}
		& 2.01 & -- \\
		\cdashline{2-4}
		& U$^{2}$Flow \settablecitationfont\clgray{CVPR'26} \cite{sun2026u}
		& 1.83 & 6.59 \\
		& U$^{2}$Flow (reproduced)
		& 1.78 & 6.58 \\
		& \textbf{TO-Flow (U$^{2}$Flow)}
		& \textbf{1.76} & \textbf{6.19} \\
		\bottomrule
\end{tabular}
\end{table}

\subsection{Unsupervised Optical Flow Estimation Results}
\label{unsur}

In unsupervised optical flow estimation, image reconstruction losses are computed from the differences between the source image and the target image warped according to the predicted flow. However, incorrect matches between visually similar locations may also produce low reconstruction errors, potentially providing misleading guidance for optical flow estimation~\cite{zhang2026learning, wang2025jasmine, liu2026disparity}. The TO-Flow framework is designed to adapt matching features to better distinguish predicted correspondences from neighboring alternatives. Therefore, TO-Flow is applied to U$^{2}$Flow, a competitive unsupervised optical flow method, to evaluate its potential benefit. The resulting variant is evaluated on the MPI Sintel and KITTI 2015 training sets and compared with the reproduced U$^{2}$Flow baseline. The originally reported baseline results and those of other representative unsupervised methods are also presented for comparison.

As shown in Tables~\ref{tb.unsupervised_sintel} and~\ref{tb.unsupervised_kitti}, TO-Flow outperforms the reproduced U$^{2}$Flow baseline on all reported metrics for both datasets. On the MPI Sintel training set, the EPE decreases from 1.59 to 1.53 pixels on the Clean pass and from 2.51 to 2.43 pixels on the Final pass. On the KITTI 2015 training set, the EPE decreases from 1.78 to 1.76 pixels, while the Fl-all error rate decreases from 6.58\% to 6.19\%. Qualitative comparisons in Fig.~\ref{unsupvised} further illustrate the improvements over the reproduced baseline. Together, these results support the effectiveness of task-oriented feature adaptation in improving optical flow accuracy under image reconstruction supervision.

\begin{table}[t]
\centering
\settablefont
\setlength{\tabcolsep}{10pt}
\caption{Component ablations of TO-Flow on the MPI Sintel and KITTI 2015 training sets. The baseline excludes all proposed components. The best results are marked in bold.}
\label{tb.component_ablation}

\begin{tabular}{@{}lcccc@{}}
		\toprule
		\multirow{2}{*}[-0.75ex]{Configurations}
		& \multicolumn{2}{c}{Sintel (train)}
		& \multicolumn{2}{c}{KITTI 2015 (train)} \\
		\cmidrule(lr){2-3}
		\cmidrule(lr){4-5}
		& Clean$\downarrow$ & Final$\downarrow$
		& EPE$\downarrow$ & Fl-all$\downarrow$ \\
		\midrule
		TO-Flow
		& \textbf{1.58} & \textbf{2.49}
		& \textbf{4.42} & \textbf{16.91} \\

		w/o channel modulation
		& 1.59 & 2.76 & 5.03 & 19.03 \\

		w/o spatial reorganization
		& 1.69 & 2.60 & 4.88 & 19.45 \\

		w/o region partition
		& 1.95 & 3.15 & 5.67 & 22.85 \\
		Baseline (WAFT-DAv2-a2)
		& 1.69 & 2.77 & 5.11 & 19.99 \\
		\bottomrule
\end{tabular}
\end{table}

\begin{table}[t]
\centering
\settablefont
\setlength{\tabcolsep}{2.67pt}
\caption{Comparison of alternative designs and feedback sources on the MPI Sintel and KITTI 2015 training sets. Ours denotes the proposed design in each group. Configuration (a4) uses ground-truth (GT) flow for feedback during both training and evaluation. The best results within each group are marked in bold, excluding (a4).}
\label{tb.design_ablation}

\begin{tabular}{@{}llcccc@{}}
		\toprule
		\multirow{2}{*}{Analysis}
		& \multirow{2}{*}{Configurations}
		& \multicolumn{2}{c}{Sintel (train)}
		& \multicolumn{2}{c}{KITTI 2015 (train)} \\
		\cmidrule(lr){3-4}
		\cmidrule(lr){5-6}
		& & Clean$\downarrow$ & Final$\downarrow$
		& EPE$\downarrow$ & Fl-all$\downarrow$ \\
		\midrule

		\multirow{4}{*}{\shortstack[l]{(a) Channel\\Modulation}}
		& (a1) SENet \settablecitationfont\clgray{CVPR'18} \cite{hu2018squeeze} & 1.82 & 2.86 & 4.91 & 19.90 \\
		& (a2) CBAM \settablecitationfont\clgray{ECCV'18} \cite{woo2018cbam} & 1.82 & 3.02 & 5.71 & 21.22 \\
		& (a3) Ours
		& \textbf{1.69} & \textbf{2.60}
		& \textbf{4.88} & \textbf{19.45} \\
		\cdashline{2-6}
		& (a4) Ours w/ GT
		& 1.66 & 2.53 & 4.47 & 17.58 \\
		\midrule

		\multirow{3}{*}{\shortstack[l]{(b) Spatial\\Reorganization}}
		& (b1) PAC \settablecitationfont\clgray{CVPR'19} \cite{su2019pixel} & 1.81 & 3.15 & 5.26 & 20.17 \\
		& (b2) Involution \settablecitationfont\clgray{CVPR'21} \cite{li2021involution} & 1.85 & 2.97 & 5.32 & 19.52 \\
		& (b3) Ours
		& \textbf{1.59} & \textbf{2.76}
		& \textbf{5.03} & \textbf{19.03} \\
		\bottomrule
\end{tabular}
\end{table}

\subsection{Ablation Studies}
\label{abl}

The first ablation study evaluates the contributions of channel modulation and spatial reorganization to TO-Flow and examines the role of region partitioning in both components.
As shown in Table~\ref{tb.component_ablation}, removing either channel modulation or spatial reorganization degrades optical flow estimation performance, supporting their complementary contributions. Both ablated variants nevertheless outperform the WAFT-DAv2-a2 baseline, demonstrating the individual effectiveness of each component.

In addition, region partitioning is removed while both adaptation components are retained. In this variant, channel modulation derives a single channel reweighting vector from matching information aggregated over the entire feature map, while spatial reorganization applies attraction and repulsion at their respective scales without distinguishing region types. This variant performs worse than the WAFT-DAv2-a2 baseline, highlighting the importance of region-dependent adaptation and supporting our analysis in Sect.~\ref{sec:introduction} that disregarding regional differences in channel contributions and spatial feature relationships can compromise flow estimation.

The second ablation study, presented in Table~\ref{tb.design_ablation}, examines whether the gains from channel modulation and spatial reorganization arise from their task-oriented designs rather than generic feature refinement along the channel and spatial dimensions, respectively. To this end, several alternative designs from prior studies are included for comparison.

The proposed channel modulation is replaced by SENet~\cite{hu2018squeeze} and CBAM~\mbox{\cite{woo2018cbam}} in two separate configurations, with spatial reorganization removed to isolate the effect of channel adaptation. Both alternatives, (a1) and (a2), yield higher optical flow estimation errors than the proposed design in (a3). These results support the importance of adapting channel contributions according to their ability to distinguish correspondences from competing candidates, beyond the information provided by within-frame feature statistics.

To further examine whether inaccuracies in the correspondence hypotheses generated by simulated flow refinement compromise the reliability of channel modulation, as discussed in Sect.~\ref{3.3}, predicted flow is replaced with ground-truth flow during both training and evaluation in (a4). This configuration yields results comparable to those of the proposed design. These results indicate that the regional aggregation and averaging across simulated iterations in Eqs.~\eqref{eq:region_matching_response}--\eqref{eq:channel_feedback} reduce reliance on individual inaccurate correspondence hypotheses.

Similarly, the proposed spatial reorganization is replaced by pixel-adaptive convolution~\cite{su2019pixel} and Involution~\cite{li2021involution} in separate configurations, with channel modulation removed to isolate the effect of spatial adaptation. Both alternatives yield higher optical flow estimation errors than the proposed design in (b3) across all four metrics. These results support our analysis that context feature adaptation benefits from distinct mechanisms for region interiors and motion boundaries, promoting consistency within coherent regions while strengthening distinctions across boundaries.

\subsection{Discussion}
\label{discus}

Despite the compelling results achieved, the proposed framework still has the following two limitations.

First, the computational efficiency of the TO-Flow framework remains limited. For example, with inputs of $436 \times 1024$ pixels, inference latency increases from 105~ms for WAFT-DAv2-a2 to 240~ms for TO-Flow (WAFT-DAv2-a2), measured on a single NVIDIA GeForce RTX 4090 GPU with a batch size of 1 and FP32 precision.
This additional computational overhead mainly stems from the five simulated refinement iterations performed by TO-Flow to assess channel matching behavior. A possible solution is therefore to explore how to integrate channel modulation into the main refinement process while efficiently reusing matching computations and preserving estimation accuracy.

Second, the applicability of TO-Flow beyond the evaluated architectures requires further validation. Although TO-Flow improves the performance of FlowSeek-T, WAFT-DAv2-a2, and U$^{2}$Flow, which collectively cover cost-volume-based and warping-based refinement mechanisms, its effectiveness in other existing model families (\eg, generative optical flow models) and in models based on emerging paradigms remains to be evaluated.

\section{Conclusion and Future Work}\label{sec:conclusion}

This work presents TO-Flow, a task-oriented feature adaptation framework for optical flow estimation. The distinct roles of matching and context features are first analyzed, revealing limitations in the representations produced by existing encoders. Based on this analysis, TO-Flow modulates matching feature channels using feedback from simulated flow refinement and spatially reorganizes context features through attraction and repulsion from coarse to fine. Extensive experiments across different optical flow architectures and datasets show that TO-Flow improves overall optical flow accuracy in both supervised and unsupervised settings and enhances cross-dataset generalization.

Future work will explore the potential of task-oriented feature adaptation for a broader range of downstream tasks that rely on accurate correspondence estimation and will also focus on reducing computational overhead to improve its practical applicability.

\small
\normalem
\bibliographystyle{IEEEtran}
\bibliography{ref}

\end{document}